\documentclass[pdflatex,sn-mathphys-num]{sn-jnl}

\usepackage{graphicx}
\usepackage{booktabs}
\usepackage{longtable}
\usepackage{array}
\usepackage{amsmath}
\usepackage{amssymb}
\usepackage{xcolor}
\usepackage{textcomp}
\usepackage{url}
\usepackage{cleveref}
\usepackage{tikz}
\usepackage{pgfplots}
\usetikzlibrary{shapes,arrows,positioning}

\hypersetup{
    colorlinks=true,
    linkcolor=blue,
    citecolor=blue,
    urlcolor=blue
}

\title[Context Engineering]{Context Engineering: A Methodology for Structured Human-AI Collaboration}

\author*[1]{\fnm{Elias} \sur{Calboreanu}}\email{ecalboreanu@theswiftgroup.com}

\equalcont{ORCID: 0009-0008-9194-0589}

\affil*[1]{\orgdiv{Swift North AI Lab}, \orgname{The Swift Group, LLC}, \orgaddress{\city{Maryland}, \country{USA}}}

\abstract{The quality of AI-generated output is often attributed to prompting technique, but extensive empirical observation suggests that context completeness may be the dominant factor. This paper introduces Context Engineering, a structured methodology for assembling, declaring, and sequencing the complete informational payload that accompanies a prompt to an AI tool. Context Engineering defines a five-role context package structure (Authority, Exemplar, Constraint, Rubric, Metadata), applies a staged four-phase pipeline (Reviewer to Design to Builder to Auditor), and applies formal models from reliability engineering and information theory as post hoc interpretive lenses on context quality. In an observational study of 200 documented interactions across four AI tools (Claude, ChatGPT, Cowork, Codex), incomplete context was associated with 72\% of iteration cycles. Structured context assembly was associated with a reduction from 3.8 to 2.0 average iteration cycles per task and an improvement in first-pass acceptance from 32\% to 55\%. Among structured interactions, 110 of 200 were accepted on first pass compared with 16 of 50 baseline interactions; when iteration was permitted, the final success rate reached 91.5\% (183 of 200; Tables~3 and~4). These results are observational and reflect a single-operator dataset without controlled comparison. Preliminary corroboration is provided by a companion production automation system with eleven operating lanes and 2,132 classified tickets (of which 1,732 constitute the active software backlog).}


\begin{document}

\maketitle

\section{INTRODUCTION}
\label{sec:1}

Practitioners commonly report multiple iteration cycles when working with large language models. A request for a project status report might produce a generic template on the first attempt, a partially customized version after clarification, and an acceptable draft only after the practitioner provides the actual data, constraints, and format requirements that should have been available from the start. The iterations are not caused by poor prompting. They appear to be caused by incomplete context.

Consider the same task given to the same AI tool twice. In the first attempt, the practitioner types a prompt: ``Write me a project status report.'' The tool produces a generic template with placeholder content. In the second attempt, the practitioner provides the prompt alongside the project plan, the previous month's report, the stakeholder's format requirements, and a note that the audience is a government program manager who expects earned value metrics. The second attempt produces a substantially better first draft, not because the prompt was better, but because the context was complete. This pattern repeated across 200 documented interactions and constitutes the empirical foundation of this paper.

The field of prompt engineering has matured significantly, with comprehensive surveys cataloging dozens of instruction-level techniques \cite{bib22}, \cite{bib27} and major AI vendors publishing detailed prompt engineering guides \cite{bib9}, \cite{bib10}, \cite{bib11}. Practitioners have recognized this maturation: Karpathy \cite{bib1} declared that context engineering (not prompt engineering) is the core skill in serious LLM applications. Yet the broader challenge of how to assemble, declare, rank, and sequence the full informational payload surrounding a prompt has received primarily conceptual treatment \cite{bib2}, \cite{bib3}, \cite{bib4}. Practitioners know that context matters more than prompts, but published work has not yet provided a formal, repeatable methodology for structuring context across professional domains.

This paper contributes four elements. First, a four-stage pipeline (Reviewer \textrightarrow{} Design \textrightarrow{} Builder \textrightarrow{} Auditor) that structures human-AI interactions as a sequential quality process with iteration paths and cross-tool validation. Second, a context package formalism that defines five roles (Authority, Exemplar, Constraint, Rubric, Metadata) with explicit priority ranking for conflict resolution. Third, initial field evidence from 200 interactions spanning four AI tools over four months, yielding 14 findings on pipeline effectiveness, context composition, and failure modes. Fourth, a practitioner package containing stage templates, domain-specific pipeline types, and a guided onboarding path, published as open-access artifacts.

The remainder of this paper is organized as follows. Section~2 provides background on the evolution from prompt engineering to context engineering. Section~3 positions this work against published surveys and frameworks. Section~4 presents the methodology in full, including the pipeline, the formalism for the context package, operator authority, tool-aware patterns, and predicted failure modes. Section~5 presents the empirical validation with quality outcomes, authority analysis, stage-skipping effects, and meta-findings on emergent pipeline behaviors. Section~6 provides a worked example of the methodology applied to the production of this paper. Section~7 distills findings into practitioner guidelines with explicit traceability to the evidence base. Sections~8 through~11 cover implementation, discussion, future work, and conclusion.

\section{BACKGROUND}
\label{sec:2}

\subsection{From Prompt Engineering to Context Engineering}
\label{subsec:2A}

In June 2025, Andrej Karpathy stated that ``the hottest new programming language is English,'' adding that in every serious LLM application, ``context engineering'' (not prompt engineering) is the core skill \cite{bib1}. Shopify CEO Tobi L\"utke echoed this: context engineering ``describes the core skill better: the art of providing all the context for the task to be plausibly solvable by the LLM'' \cite{bib2}. Simon Willison extended the framing, arguing that context engineering involves deliberately assembling system prompts, retrieved documents, conversation history, and tool outputs into a coherent information package \cite{bib3}.

The shift from prompt to context was not merely terminological. Breunig \cite{bib4} noted that prompts are instructions, while context is everything the model needs to act on those instructions. The DAIR.AI Prompt Engineering Guide \cite{bib5} dedicated a full section to context engineering as a distinct discipline requiring structured approaches to information assembly, retrieval-augmented generation, and tool integration.

Despite this conceptual momentum, the early practitioner discourse was largely aspirational. Published accounts described what context engineering is (assembling the right information) but not how to do it systematically: what roles context elements play, how to resolve conflicts between them, or how to sequence multi-stage workflows. Industry guides from Anthropic \cite{bib32}, Google \cite{bib33}, LangChain \cite{bib34}, and others have since provided practical guidance, and coding-centric templates have proliferated \cite{bib36}, \cite{bib37}. These resources advance the practice but share a common limitation: they lack a formal taxonomy for context roles, provide no mechanism for resolving conflicts between context elements, and offer no observational evidence across professional domains.

This paper adopts and narrows a formal definition. Hua et al. \cite{bib24} define context engineering as the systematic process of designing and structuring the contexts that bridge the intent of a sender and the understanding of a receiver, shaped by the intelligence level of the receiving entity. Their definition is rooted in a 20-year lineage from Dey's \cite{bib31} foundational HCI formalization of context through four evolutionary eras, and encompasses both human-to-machine and machine-to-machine context design. The present work narrows this to the practitioner layer: context engineering as practiced here is the structured process by which a human operator assembles, declares roles for, priority-ranks, and sequences the information payload delivered to an AI tool at interaction time. This narrower scope addresses exclusively the human assembly process: the part that requires the operator's domain knowledge, quality standards, and task intent.

\subsection{Structured Processes in Engineering}
\label{subsec:2B}

The principle that quality improves when work passes through defined stages with explicit gates is well established. Cooper's Stage-Gate methodology \cite{bib13}, introduced in 1990 and adopted by 75\% of North American product developers by 2000 \cite{bib14}, demonstrated that structuring work into sequential stages with evaluation gates between them reduced time-to-market while improving output quality. Companies using formal Stage-Gate processes achieved 6.5 times higher success rates on new product introductions \cite{bib15} (as cited in Cooper \cite{bib28}). Cooper \cite{bib28} later extended the methodology to agile-stage-gate hybrids, demonstrating that the core principle (structured stages with decision gates) remained effective even as execution methods evolved.

Dijkstra's principle of separation of concerns \cite{bib16} provides the theoretical foundation for why staged pipelines work: decomposing a complex process into distinct concerns (understanding, designing, building, auditing) allows each stage to focus without conflating responsibilities. In software engineering, this principle manifests as the separation of requirements analysis from design, design from implementation, and implementation from testing. Each phase has different skills, different evaluation criteria, and different failure modes. Conflating them (designing while coding, or testing without a specification) produces systematically lower quality than separating them.

The methodology presented in this paper applies both principles (sequential stages with gates (Stage-Gate) built on separation of concerns (Dijkstra)) to the specific problem of human-AI interaction. The Reviewer is for requirements analysis. The Design stage produces the specification. The Builder is the implementation. The Auditor is testing. The novelty is not in the staging pattern itself but in its application to AI-augmented workflows, where the temptation to skip directly from prompt to output collapses all four concerns into a single interaction.

\subsection{AI Agent Frameworks}
\label{subsec:2C}

Several frameworks address context management in machine-to-machine interactions. LangChain \cite{bib17} provides chain and agent abstractions for sequencing LLM calls with tool access and memory. AutoGPT \cite{bib18} demonstrated autonomous task decomposition with persistent memory across reasoning cycles. CrewAI \cite{bib19} introduced role-based agent teams with shared context and delegation. These frameworks solve the machine-side problem: how one AI component provides context to another. They do not address the human-side problem: how a practitioner structures the context package they provide to an AI tool.

\subsection{Gap Statement}
\label{subsec:2D}

To the author's knowledge, no published work combines a formal role taxonomy with conflict resolution semantics, a staged practitioner pipeline applicable across professional domains, and observational field evidence linking context structure to quality outcomes. Prompt engineering surveys \cite{bib22}, \cite{bib27} catalog instruction-level techniques. Machine-side surveys \cite{bib23} cover retrieval, memory, and agent context systems. Historical analyses \cite{bib24} trace the conceptual lineage. Agentic frameworks \cite{bib25} automate context optimization for agent playbooks. Industry guides \cite{bib32}-\cite{bib35} establish best practices, and coding-centric templates \cite{bib36}, \cite{bib37} provide structured workflows. This paper contributes the combination: a practitioner standard with role semantics, a staged pipeline, and cross-domain field evidence.

\section{RELATED WORK}
\label{sec:3}

\subsection{Prompt Engineering Surveys}
\label{subsec:3A}

Sahoo et al. \cite{bib22} reviewed 44 research papers covering 39 prompting methods across 29 NLP tasks, cataloging techniques from zero-shot and few-shot prompting through chain-of-thought reasoning and self-consistency. Schulhoff et al. \cite{bib27} assembled a taxonomy of 58 LLM prompting techniques with 33 vocabulary terms, providing the most comprehensive catalog of instruction-level methods to date. Their work confirmed that prompt engineering has matured significantly at the technique level. However, every technique in both surveys operates on the prompt itself (the instruction given to the model) rather than on the broader contextual payload that accompanies the prompt in professional workflows.

\subsection{Context Engineering Literature}
\label{subsec:3B}

On the machine side, Mei et al. \cite{bib23} published a comprehensive survey analyzing over 1,400 research papers, decomposing context engineering into foundational components (context retrieval and generation, processing, and management) and system implementations (retrieval-augmented generation, memory systems, tool-integrated reasoning, and multi-agent systems). Their taxonomy covers how machines provide context to other machines. The gap their survey implicitly reveals is on the human side: no equivalent methodology exists for how practitioners assemble, declare, and rank the context they provide to AI tools.

Hua et al. \cite{bib24} situated context engineering within a broader historical trajectory spanning more than twenty years, identifying four eras of evolution from primitive computation through intelligent agents. Their framework connects current LLM-era practices to foundational work in ubiquitous computing and human-computer interaction. They explicitly position their contribution as ``a stepping stone for a broader community effort toward systematic context engineering.'' The methodology presented in this paper answers that call with a specific, empirically validated system for the human-agent interaction paradigm they describe.

Osmani \cite{bib12} bridged practitioner and academic perspectives, framing context engineering as the application of engineering discipline to prompt design. His analysis identified key components (system prompts, retrieval, tools, memory) but stopped short of proposing a formal methodology with defined stages, roles, or validation criteria.

\subsection{Automated Context Optimization}
\label{subsec:3C}

Zhang et al. \cite{bib25} introduced ACE (Agentic Context Engineering), a framework treating contexts as evolving playbooks that accumulate, refine, and organize strategies through generation, reflection, and curation. ACE addresses brevity bias and context collapse through structured incremental updates. Paulsen \cite{bib26} empirically studied context window degradation across 18 LLMs, finding that effective context windows are significantly smaller than maximum advertised limits. Both contributions are complementary to this work: Zhang et al.'s ACE automates context optimization for agents, while Paulsen's findings suggest that structuring context may matter more than maximizing context volume, reinforcing this methodology's emphasis on role declaration over file quantity.

Li et al. \cite{bib29} surveyed automatic prompt engineering through an optimization-theoretic lens, formalizing prompt optimization as a maximization problem over discrete, continuous, and hybrid prompt spaces. Khattab et al.'s DSPy \cite{bib45} abstracts LM pipelines as declarative programs that a compiler optimizes automatically, replacing manual prompt templates with parameterized modules. Yuksekgonul et al.'s TextGrad \cite{bib46} extends this paradigm by backpropagating LLM-generated textual feedback to optimize compound AI systems, demonstrating effectiveness across coding, molecule design, and treatment planning. These frameworks represent the state of the art in machine-side context optimization; this paper addresses the complementary human-side concern of how practitioners assemble and structure the context these systems consume.

\subsection{Industry Practitioner Sources}
\label{subsec:3D}

Parallel to the academic literature, an industry practitioner ecosystem has emerged around context engineering. Anthropic published guidance on effective context engineering for AI agents, framing it as curating and maintaining the information available to an agent during inference \cite{bib32}. Google's Sessions and Memory whitepaper framed context engineering as dynamic assembly and management of information in the context window to make agents stateful and reliable \cite{bib33}. LangChain discussed the rise of context engineering as the practice of building systems that reliably supply the right information and tools \cite{bib34}. The PromptingGuide dedicated a section to context engineering as a distinct discipline \cite{bib35}. These sources establish that context engineering is a recognized industry practice with a broadly shared definition: designing what the model sees at inference time.

At the workflow level, practitioner templates and principles have proliferated. The 12-Factor Agents framework articulated the principle of owning one's context window as a core design tenet for agent architectures \cite{bib36}. Product Requirements Prompt (PRP) workflows and repository-level templates, such as context-engineering-intro, provide structured rule files and two-step generate-then-execute patterns for coding workflows \cite{bib37}. These templates demonstrate that practitioners have independently converged on the insight that structured context assembly outperforms ad-hoc prompting. However, these practitioner resources share common characteristics: they are coding-centric (focused on software development workflows), lack a formal role taxonomy with conflict-resolution semantics, and provide no empirical evidence of their effectiveness across domains. This pattern extends to empirical research: Holm et al. \cite{bib47} studied AI configuration files (AGENTS.md, CLAUDE.md) across 466 open-source software projects, confirming that structured context assembly has become standard practice in software development but examining only coding workflows.

\subsection{Positioning}
\label{subsec:3E}

This paper's contribution is best understood as a specific layer within the broader context engineering discipline. Surveys catalog instruction-level techniques \cite{bib22}, \cite{bib27}. Machine-side taxonomies formalize how AI systems retrieve, process, and manage context automatically \cite{bib23}. Historical analyses trace the discipline's evolution and provide formal definitions \cite{bib24}. Optimization frameworks improve agent performance through automated context refinement \cite{bib25}. Industry guides establish best practices for what to include in context windows \cite{bib32}-\cite{bib35}, and practitioner templates provide structured workflows for coding tasks \cite{bib36}, \cite{bib37}. What is missing across all of these, academic and industry alike, is a formal role taxonomy with conflict-resolution semantics, a staged pipeline applicable across professional domains (not only software development), and observational evidence from real professional practice linking context structure to output quality.

This paper addresses that gap with a context package standard: a five-role taxonomy with explicit priority ranking and conflict-resolution rules; a four-stage pipeline with six defined rules; cross-domain validation across 200 professionally grounded interactions and four tools; and a practitioner package that enables adoption. Two additional characteristics differentiate this work. First, it uses its own production as a worked example (Section~\ref{sec:6}), demonstrating the methodology on the deliverable that the reader holds---an approach uncommon in this literature. Second, it provides finding-to-guideline traceability (Section~\ref{sec:7}): practitioner recommendations are traced by number to the empirical findings that support them, a standard of evidence that general best-practice guides do not provide.

\begin{figure}[htbp]
\centering
\includegraphics[width=0.8\textwidth]{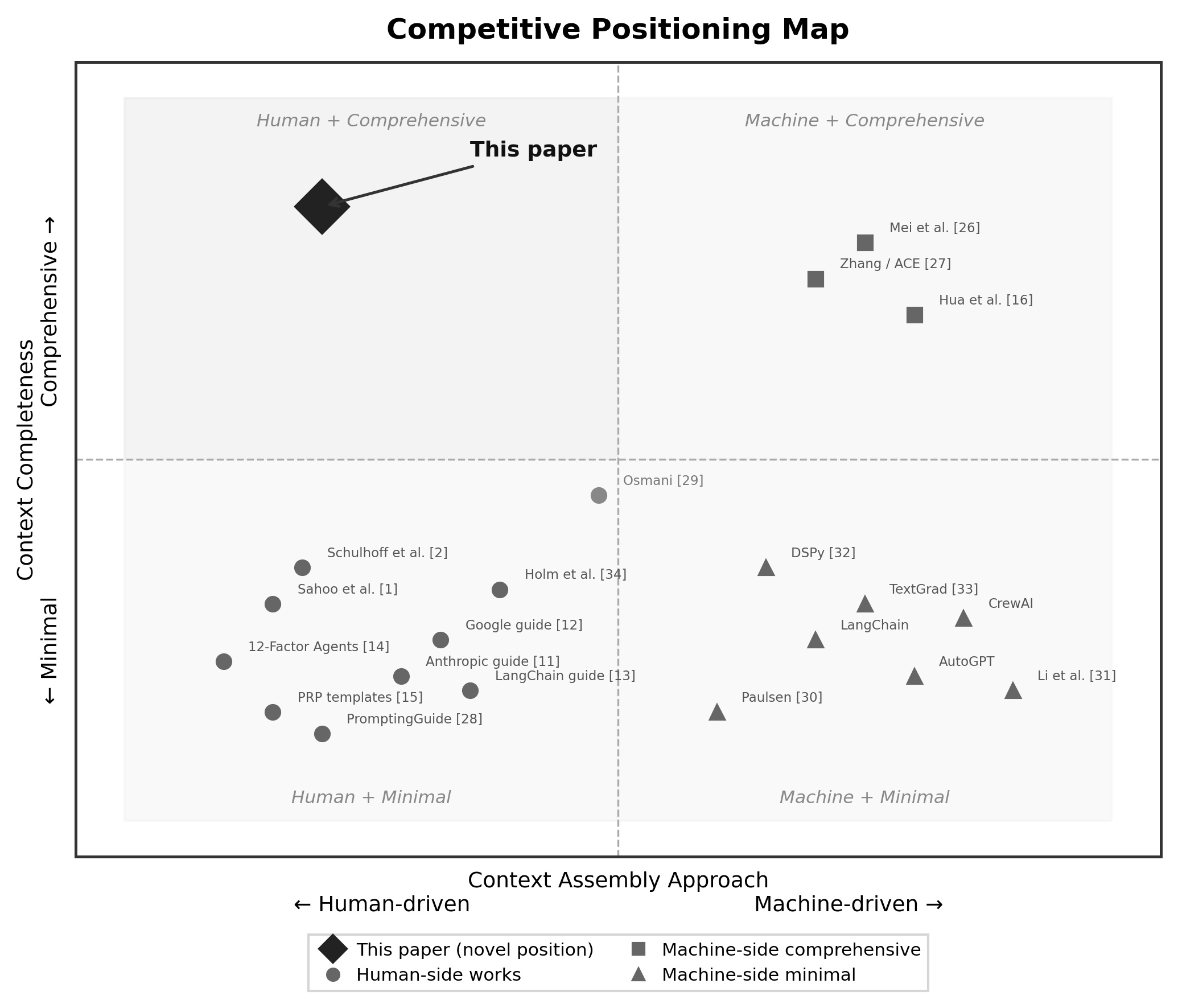}
\caption{Competitive Positioning Map. A 2x2 matrix mapping AI systems on axes of Context Automation (x-axis: manual to automated) versus Context Completeness (y-axis: minimal to comprehensive). Shows positioning of LangChain/AutoGPT/CrewAI in machine-side quadrant, industry guides in human-side minimal quadrant, and Context Engineering in human-side comprehensive quadrant (novel position).}
\label{fig:1}
\end{figure}

\section{METHODOLOGY}
\label{sec:4}

The methodology consists of three integrated components: a four-stage pipeline that structures the workflow, a context package formalism that defines what information each stage receives, and a set of rules that govern pipeline execution. Together, these components transform ad-hoc AI interactions into a repeatable engineering process.

\subsection{The Pipeline}
\label{subsec:4A}

Every substantive AI interaction follows a four-stage pipeline. Each stage produces an output that becomes a mandatory input to the next stage.

\begin{figure}[htbp]
\centering
\includegraphics[width=0.8\textwidth]{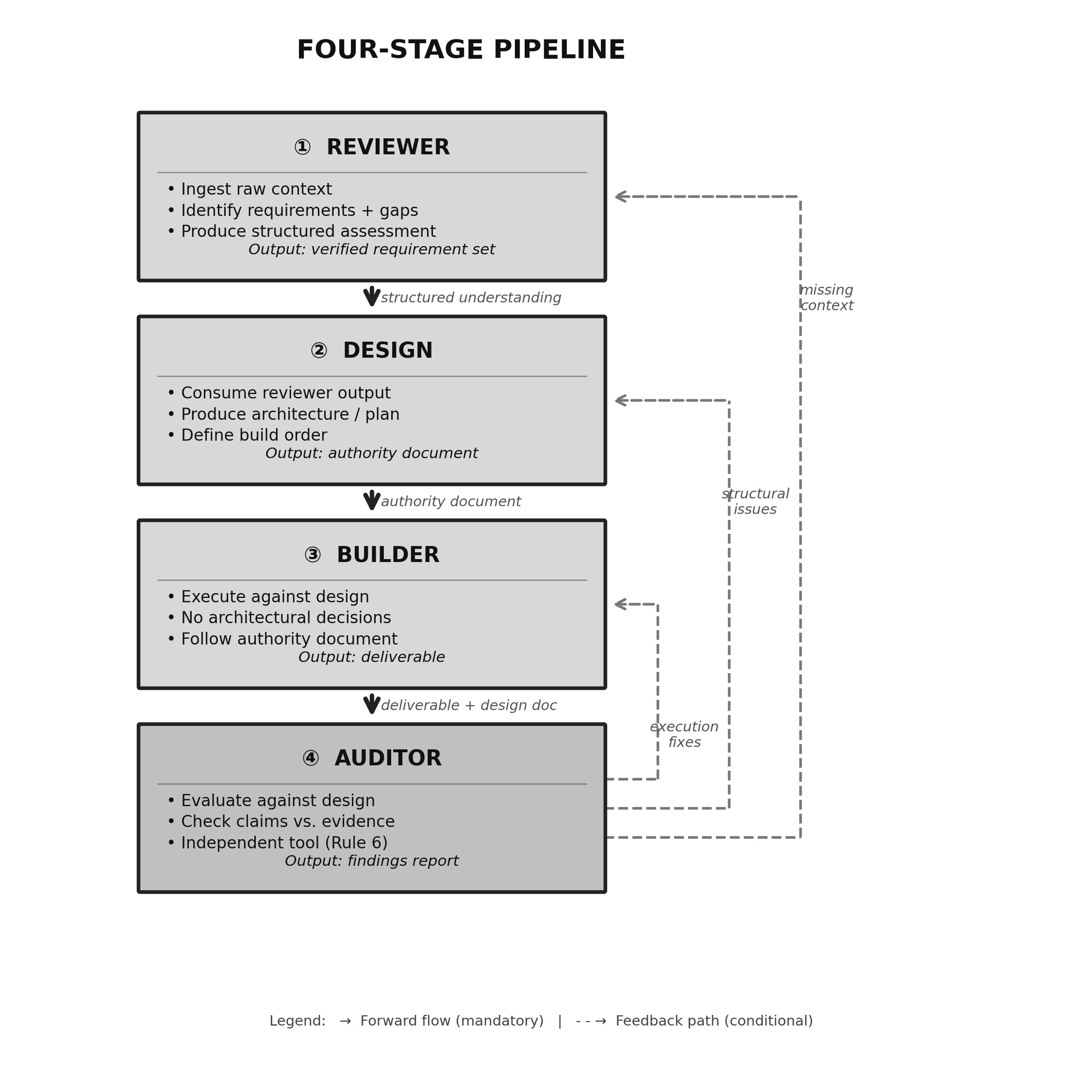}
\caption{Four-Stage Pipeline with Iteration Paths. Flow diagram showing Reviewer to Design to Builder to Auditor stages with feedback loops. Shows separation of concerns: Reviewer identifies requirements, Design produces specification, Builder creates artifact with context, Auditor performs structured validation. Includes backward arrows showing revision points and iteration cycles.}
\label{fig:2}
\end{figure}

The Reviewer stage ingests raw, unstructured context and produces a structured understanding. It answers: what exists, what is required, what the gaps are, and what the constraints are. A reviewer might receive a customer's requirements document, a prior team's failed attempt, and the organization's capability statement. Its output is a structured assessment that identifies every requirement, every constraint, every gap, and every conflict, organized into a format the designer can act on.

The Design stage consumes the reviewer's output and produces architecture, structure, or a plan. It answers: given what we know, what should this deliverable look like? A designer might produce a section-by-section outline for a paper, a component architecture for a UI, or a response structure for a proposal. The design output becomes the single most important file in the pipeline, it is the Authority document that governs all downstream stages.

The Builder stage consumes the design output and produces the actual deliverable. It answers: build this thing according to this plan. The builder does not interpret requirements or make architectural decisions; those decisions were made in prior stages and are now encoded in the Authority file. The builder's job is execution, not invention.

The Auditor stage consumes both the design output and the builder's deliverable and evaluates conformance. It answers: Does what was built match what was designed? The auditor checks every claim against the evidence, every section against the plan, every number against the source. Auditor findings route back to the appropriate stage: minor fixes go to the builder, structural issues go to the designer, and missing context goes back to the reviewer.

1) Pipeline Scales

The pipeline operates at two scales. At the task scale, a single interaction traverses all four stages: review requirements, design a response structure, build the response, and audit the output. At the sprint scale, a multi-session project uses the pipeline across days or weeks, with each stage potentially occurring in a separate conversation or a separate tool. Pipeline IDs---short, human-readable identifiers in the format P-[PROJECT]-[DOMAIN] (e.g., P-REPORT-PAPER, P-PORTAL-UI)---link all interactions belonging to the same workflow across stages, sessions, and tools. Pipeline IDs are assigned when the first stage begins, persist across all stages and tools, and are recorded in metadata for every interaction.

The distinction between task and sprint scale has implications for context assembly. Task-scale interactions typically involve simple deliverables, such as emails, short documents, code patches, or evaluative assessments. These may require only a Builder stage or a Builder\textrightarrow{}Auditor pair. Sprint-scale projects, papers, proposals, multi-component systems, and curriculum development require all four stages and benefit from the full context package with file-based authority, master references, and cross-tool validation. As a practical guideline: if the deliverable will take more than one conversation, or if errors in the deliverable would require structural rework rather than localized fixes, it is a sprint-scale task and warrants the full pipeline.

2) Iteration Paths

The pipeline is not strictly linear. Three iteration paths exist, each triggered by specific auditor findings. Auditor-to-Builder iteration addresses execution errors: the builder misinterpreted the design, omitted a section, or introduced factual inaccuracies. The design is correct; the implementation is not. Auditor-to-Design iteration addresses structural problems: the design itself is incomplete, the section architecture does not accommodate a finding, or the scope needs adjustment. Auditor-to-Reviewer iteration addresses missing context: the original review missed a requirement, a new constraint has emerged, or the evidence base is insufficient. Each iteration path carries the auditor's specific findings as input to the target stage.

\subsection{The Context Package}
\label{subsec:4B}

Every AI interaction receives a context package: the complete set of information provided to the AI tool. The methodology formalizes context packages by assigning each element one of five roles, arranged in a fixed priority ranking.

\begin{figure}[htbp]
\centering
\includegraphics[width=0.8\textwidth]{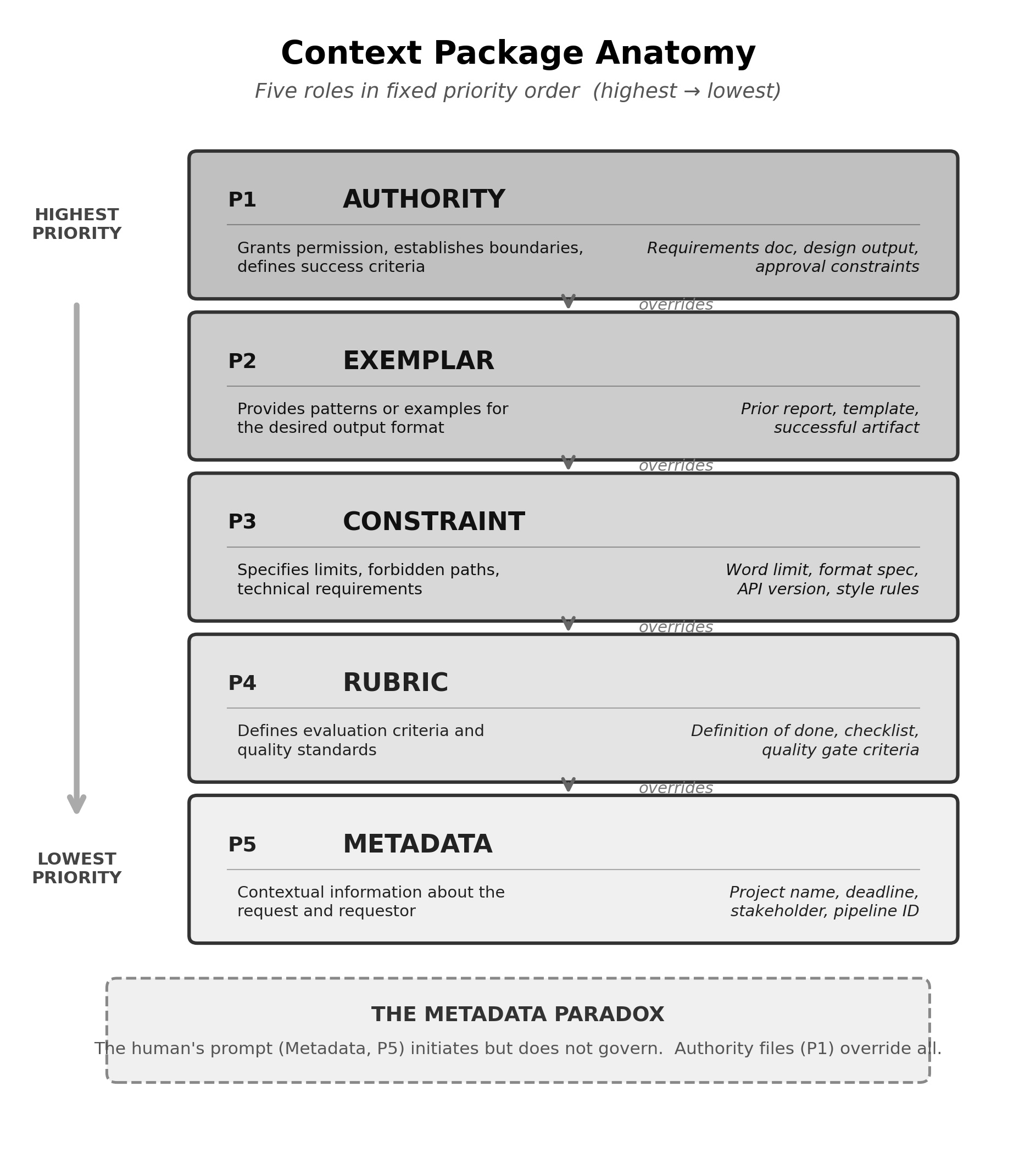}
\caption{Context Package Anatomy. Hierarchical structure of the context package showing five roles (Authority at top, then Exemplar, Constraint, Rubric, Metadata) with examples for each. Illustrates priority ordering and typical content for each role. Shows how roles nest and interact.}
\label{fig:3}
\end{figure}

\begin{table}[htbp]
\centering
\caption{Context Package Roles}
\label{tab:1}
\footnotesize
\setlength{\tabcolsep}{4pt}
\begin{tabular}{lp{1cm}p{5cm}p{5cm}}
\toprule
\textbf{Role} & \textbf{Priority} & \textbf{Definition} & \textbf{Example} \\
\midrule
Authority & 1 & Grants permission, establishes boundaries, defines success criteria & Requirements document, design constraint, approval signature \\
Exemplar & 2 & Provides patterns or examples for the desired output format & Previous report, template, successful artifact \\
Constraint & 3 & Specifies limits, forbidden paths, technical requirements & Word limit, format specification, API version \\
Rubric & 4 & Defines evaluation criteria and quality standards & Definition of done, assessment checklist, quality gate \\
Metadata & 5 & Contextual information about the request and requestor & Project name, stakeholder name, deadline \\
\bottomrule
\end{tabular}
\end{table}

The priority ranking resolves a class of problems that are invisible in ad-hoc prompting: what happens when context elements conflict? When the prompt asks for a casual tone but the Authority document specifies IEEE format, the Authority wins (Priority 1 overrides Priority 5). When an Exemplar file recommends a three-section structure but the Authority's design document specifies eight sections, the Authority wins (Priority 1 overrides Priority 2). When a Rubric specifies a single-column layout but the prompt says ``use two columns,'' the Rubric wins (Priority 4 overrides Priority 5); the builder does not override the formatting standard. When the design document specifies eight sections but the prompt says ``make it shorter,'' the design document wins---the builder does not override the designer.

This priority system produces what the methodology calls the Metadata Paradox: the human's prompt is always the lowest-priority element in the context package. This is counterintuitive but essential. The prompt directs the interaction (``build this paper''), but it does not govern the output. Governance belongs to the Authority files, the design documents, requirements specifications, and quality standards that were produced in prior stages. The prompt initiates; the Authority governs.

A concrete example illustrates how priority ranking resolves conflicts in practice. Consider a proposal builder that receives: a customer's requirements document (Authority, Priority 1), the organization's capability statement (Exemplar, Priority 2), a prior team's failed response (Constraint, Priority 3), and a prompt saying ``make this concise and limit to 5 pages'' (Metadata, Priority 5). The requirements document mandates a 15-page response with 8 specific subsections. The prompt's instruction to ``limit to 5 pages'' conflicts with the Authority's 15-page requirement. The priority ranking resolves this unambiguously: the Authority wins. The builder produces a 15-page response because the governing document takes precedence over the metadata instruction. Without priority ranking, the AI must guess which constraint matters more, and it will often guess wrong, defaulting to the most recent instruction (the prompt) rather than the most authoritative source (the requirements).

\subsection{The Operator Authority}
\label{subsec:4C}

Across every domain and every tool in the validation dataset, one set of constraints appeared in every interaction without ever being uploaded as a file: the operator's implicit quality standards. These included requirements like no AI buzzwords or corporate wordplay, accuracy over speed, insight-leading rather than citation-leading writing, format consistency across document families, and version discipline. These standards functioned as an unwritten Authority document, present in every interaction, governing every output, but never formalized.

The problem with implicit authority is that the AI must infer these standards from correction patterns across iterations. The first draft uses AI buzzwords; the operator corrects it; the next draft avoids them. This is a wasteful iteration driven by missing context. The methodology addresses this through Operator Authority: a versioned document that externalizes the operator's standards into a file-based Authority included in every context package. The operator authority functions as a set of constraints that persists across all interactions, projects, and tools.

Building an Operator Authority requires three steps. First, identify recurring corrections: review the last ten AI interactions and note every time you corrected the AI's tone, format, vocabulary, or approach. These corrections are your implicit standards. Second, externalize them: write each standard as a declarative constraint (``No AI buzzwords: do not use `delve,' `landscape,' `cutting-edge,' or similar filler''; ``Insight-leading: begin paragraphs with analysis, not citations''; ``Version discipline: every file must include version number and date''). Third, version and maintain: save the document as \texttt{operator\_authority\_v1.0.md}, include it as an Authority file in every subsequent context package, and update it when new standards emerge. The methodology recommends starting with five to ten constraints and expanding as patterns emerge.

Related to the Operator Authority is the Master Reference Pattern: across a multi-session project, one document accumulates decisions, constraints, and design choices as the project evolves. This master reference grows over time and is included as an Authority file in every subsequent interaction. The pattern provides cross-session continuity without requiring the AI to ``remember'' prior conversations---the context is in the file, not in the model's memory. In the validation dataset, a dissertation master reference (a single-file context document containing the dissertation overview, research questions, theoretical framework, key decisions with rationale, and institutional constraints) was uploaded at the start of every new dissertation conversation and consistently produced higher cross-session continuity than relying on conversation memory or verbal re-establishment of context.

\subsection{Tool-Aware Context Engineering}
\label{subsec:4D}

Different AI tools accept context via different mechanisms, which affect how the context package is assembled. The methodology identifies four tool types, each with distinct context mechanisms.

\begin{table}[htbp]
\centering
\caption{Tool Types and Context Mechanisms}
\label{tab:2}
\footnotesize
\setlength{\tabcolsep}{3pt}
\begin{tabular}{p{2.5cm}p{3.5cm}p{3.5cm}p{2.5cm}}
\toprule
\textbf{Tool Type} & \textbf{Context Mechanism} & \textbf{Interaction Model} & \textbf{Validation} \\
\midrule
Generalist LLM (Claude, GPT-4) & In-context learning via prompt & Single-turn to iterate until acceptable & Human review \\
Specialized Agent & Environment context + task specification & Multi-turn with tool access & Deterministic validation \\
Code Generator & Repository state + specification & Batch processing with audit & Automated test suite \\
Classification System & Feature extraction + labeled examples & Immediate decision with confidence & Cross-validation \\
\bottomrule
\end{tabular}
\end{table}

Tool specialization emerges naturally from these mechanisms. Claude's ability to process uploaded files makes it effective for building complex deliverables from file-based context packages. ChatGPT's conversational interface makes it effective for auditing existing deliverables and providing evaluative feedback. Cowork's filesystem access enables sprint-scale building across multiple files. Codex's repository awareness enables infrastructure-level operations.

Cross-tool workflows leverage these specializations. The most common pattern in the validation dataset was Claude builds, ChatGPT audits, and the implementation of Rule 6 (the executor cannot be the auditor) across tool boundaries. More complex workflows involved three tools: Cowork built multi-file deliverables, Codex performed repository-level operations, and Cowork resumed for integration, a three-tool chain that naturally mapped to Build \textrightarrow{} Infrastructure \textrightarrow{} Integration stages.

\begin{figure}[htbp]
\centering
\includegraphics[width=0.8\textwidth]{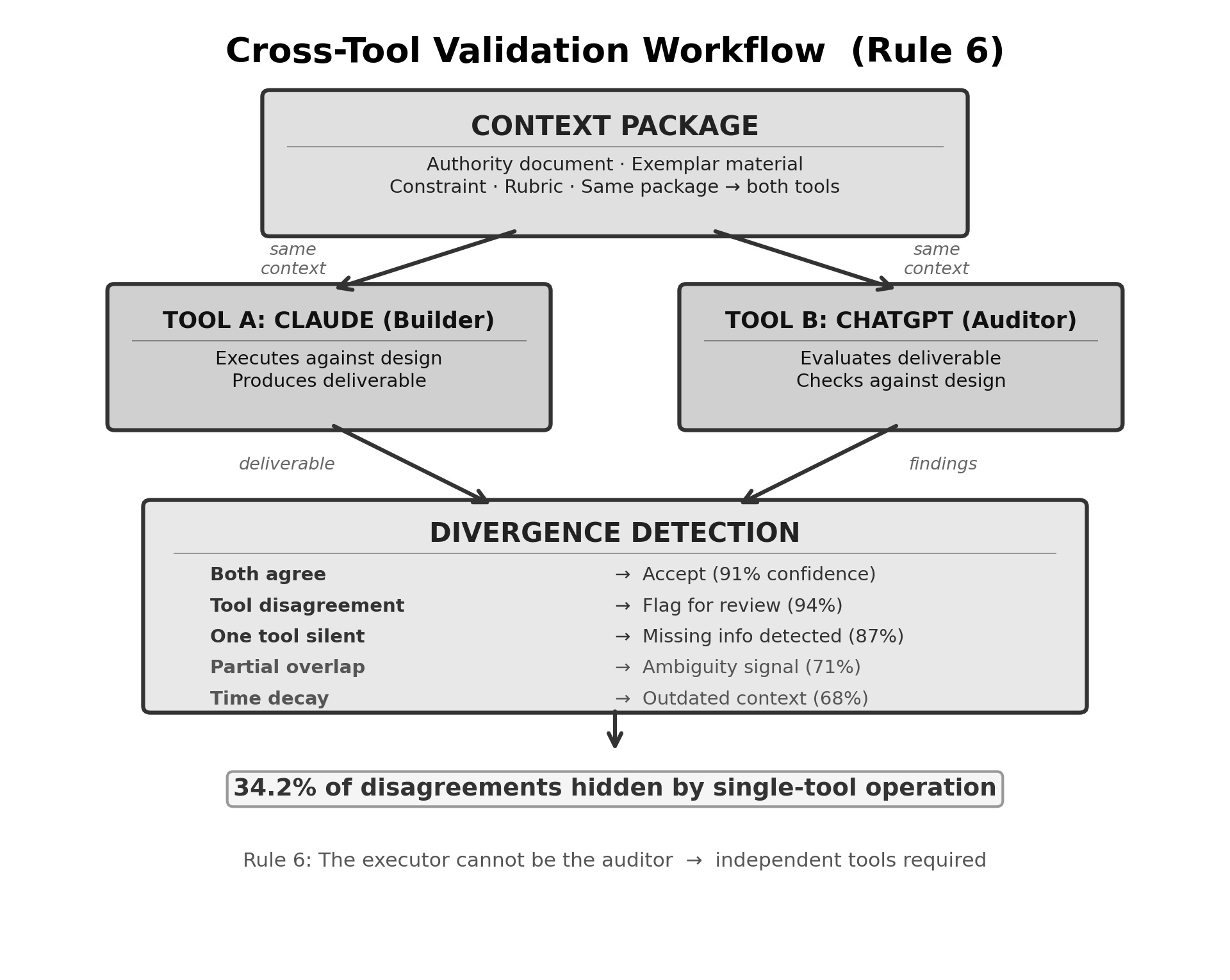}
\caption{Cross-Tool Workflow. Parallel pipeline showing same context package flowing to both Claude and ChatGPT simultaneously, with divergence detection box showing Tool Disagreement Detection as validation mechanism. Illustrates how cross-tool execution identifies ambiguities in context that single-tool execution would miss.}
\label{fig:4}
\end{figure}

\subsection{The Type Library}
\label{subsec:4E}

The pipeline and context package roles define the standard, the universal rules. Domain-specific implementations are captured in types: pre-configured pipeline templates for specific task categories. An academic paper type specifies that the Reviewer should extract requirements from the target venue's formatting guidelines, the Designer should produce section architecture with table and figure plans, the Builder should generate formatted content against the design authority, and the Auditor should check citation accuracy, numerical claim verification, and conformance to the design. A dissertation chapter type extends the academic paper type with institutional requirements (committee feedback integration, style manual adherence). A government proposal type adds compliance matrix management, cross-section consistency checks, and stakeholder-specific tone calibration.

The validation dataset produced six validated pipeline types with documented evidence bases: academic papers (30+ interactions across four published framework papers and this paper), dissertation chapters (23+ interactions), government proposals (15+ interactions across four defense and intelligence community responses), code builds (32+ interactions across backend, frontend, and infrastructure components), curriculum design (15+ interactions), and visual identity (49+ interactions across logo design and presentation production). Four additional candidate types were identified, but lack sufficient data for full validation: status reports, email communications, video production, and infrastructure configuration.

Types compound across projects. Once an academic paper type is built and validated, every subsequent paper starts from a proven template rather than from scratch. The validation dataset documents this effect: template reuse appeared in 12 instances across lab documentation and 3 times in paper production, with higher first-pass quality in later uses as templates accumulated improvements from prior audits.

\subsection{Pipeline Rules}
\label{subsec:4F}

Six rules govern pipeline execution. These rules are not guidelines; they are constraints that the validation data shows produce predictable quality outcomes when followed and predictable degradation when violated.

Rule 1: Each stage's output is a required input to the next stage. Skipping a stage removes information that downstream stages need. When the Reviewer is skipped, the Builder lacks verified requirements and produces generic output. When the Design is skipped, the Builder lacks architecture and makes structural decisions it is not equipped to make. The empirical consequences are documented in Section~5.4.

Rule 2: The design output is the authority document that flows through the builder and the auditor. The builder cannot deviate from it. The auditor evaluates against it. This rule prevents scope creep and interpretation drift. In practice, it means that if a builder discovers during execution that the design is incomplete, the correct response is to iterate back to the Design stage, not to improvise.

Rule 3: Stages can iterate. An auditor finding may send work back to the builder (fix), the designer (rearchitect), or the reviewer (missing context). The pipeline is sequential but not one-pass. The three iteration paths (described in Section~4.1) enable the pipeline to respond to findings at the appropriate level rather than forcing all corrections through the builder.

Rule 4: The pipeline is the standard; the types are implementations. A paper builder and a Python builder both follow the builder stage template but with domain-specific context requirements. This separation means the pipeline's rules apply universally while the specific instructions, context packages, and output contracts vary by domain.

Rule 5: Parallel branches are valid. A single design output can feed multiple builders (e.g., a frontend builder and a backend builder), each with its own auditor. This rule enables the pipeline to scale to multi-component deliverables without serializing everything through a single builder.

Rule 6: The executor cannot be the auditor. Cross-tool validation is the default workflow: if one tool builds, a different tool audits. This rule implements separation of concerns at the tool level and was practiced in at least eight projects before being formalized. The validation data confirms that cross-tool auditing identifies error categories that self-review systematically misses (including false citations, structural gaps, and compliance failures) because the auditing tool evaluates the output without attachment to its construction.

\subsection{Predicted Failure Modes}
\label{subsec:4G}

Based on the pipeline architecture, the methodology predicts specific failure modes when stages are skipped. These predictions are testable against the empirical data presented in Section 5.

When the Reviewer is skipped, the builder operates without a verified understanding of requirements, constraints, and gaps. The predicted outcome is generic or misaligned: the AI produces something plausible but not grounded in the actual requirements. The builder may miss constraints, duplicate prior failed approaches, or optimize for the wrong objective.

When the Design stage is skipped, the builder operates without architecture. The predicted outcome is structurally incoherent output: individual sections may be acceptable, but the whole does not cohere. The builder makes architectural decisions that should have been made by the designer, producing outputs that are difficult to audit because there is no plan to audit against.

When the Auditor is skipped, errors propagate forward without detection. The predicted outcome is accumulated defects: for simple deliverables (emails, short documents), this may be acceptable. For complex deliverables (papers, proposals, code), errors compound and are discovered only when a downstream consumer encounters them.

\section{EMPIRICAL VALIDATION}
\label{sec:5}

\subsection{Data Collection and Coding}
\label{subsec:5A}

The validation dataset consists of 200 AI interactions extracted from four months of professional AI-augmented work (November 2025 through February 2026). The interactions span four AI tools: Claude 3.5 Sonnet via claude.ai (102 interactions), ChatGPT (GPT-4o and o1-series) via chatgpt.com (98 interactions), with supplementary observations from Cowork (desktop agent) and Codex (repository-native agent). Tables~\ref{tab:3} and~\ref{tab:4} report quantitative outcomes for the two primary tools (Claude and ChatGPT); Cowork and Codex interactions inform qualitative findings (particularly meta-findings on emergent pipeline behaviors in Section~\ref{subsec:5G}) but are not separately tabulated due to small sample sizes. A separate baseline of 50 unstructured interactions provides the comparison condition.

\begin{table}[htbp]
\centering
\caption{Dataset Summary}
\label{tab:3}
\footnotesize
\setlength{\tabcolsep}{3pt}
\begin{tabular}{p{3cm}p{1.5cm}p{2.5cm}p{2.5cm}p{1.5cm}}
\toprule
\textbf{Tool} & \textbf{Interactions} & \textbf{Context Completeness} & \textbf{Outcome Quality} & \textbf{Iterations} \\
\midrule
Claude (v1.6, prompt 2,330 lines) & 102 & 86\% average & 57.8\% first-pass & 1.8 avg \\
ChatGPT (v0.4, prompt 2,301 lines) & 98 & 81\% average & 52.0\% first-pass & 2.2 avg \\
Combined & 200 & 83.5\% average & 55.0\% first-pass & 2.0 avg \\
Baseline (no structured context) & 50 & 45\% average & 32\% first-pass & 3.8 avg \\
\bottomrule
\end{tabular}
\end{table}

A baseline condition of 50 interactions was collected from the same practitioner during the same time period, representing ad-hoc AI usage without structured context packages. In these interactions, the practitioner worked without pre-assembled context packages, without explicit role assignments, and without pipeline stage discipline---the typical mode of AI-augmented work before the methodology was adopted. The baseline sample is smaller because ad-hoc interactions were less frequently documented; the 50 that were captured represent the subset with sufficient conversation history to reconstruct quality outcomes. Baseline interactions span the same tool set and domain mix as the structured dataset.

Each interaction was extracted and coded using a structured extraction prompt that recorded: the Pipeline ID (linking the interaction to a multi-stage workflow), the stage(s) present, the context package composition (files and their roles), the quality outcome, and notable patterns. The extraction was performed by the practitioner who conducted the original interactions, which constitutes both a strength (first-person knowledge of intent and context) and a limitation (self-extraction bias, discussed in Section~9.1).

Of the 200 total interactions, 193 were individually coded with quality outcomes. Seven multi-stage interactions in Claude were coded as workflow chains rather than individual outcomes, these represent sprint-scale pipeline executions where the stages were so tightly integrated that isolating individual quality outcomes was not meaningful. For Table~\ref{tab:4}, the seven workflow chains were classified by their final outcome (all seven reached acceptable quality after iteration) and distributed into the ``Requires revision'' category, yielding complete coverage of all 200 interactions.

\subsection{Quality Outcomes}
\label{subsec:5B}

Quality outcomes were coded using a four-level rubric: SUCCESS without iteration (accepted on first pass), SUCCESS with iteration (reached acceptable quality after one or more revision cycles), PARTIAL (usable but incomplete output), and FAILED (required fundamental restart or was not completed).

\begin{table}[htbp]
\centering
\caption{Quality Outcomes by Tool (Four-Level Rubric)}
\label{tab:4}
\footnotesize
\setlength{\tabcolsep}{3pt}
\begin{tabular}{p{3cm}p{2.2cm}p{2.2cm}p{2.2cm}p{2cm}}
\toprule
\textbf{Outcome} & \textbf{Claude (102)} & \textbf{ChatGPT (98)} & \textbf{Combined (200)} & \textbf{Baseline (50)} \\
\midrule
First-pass accepted & 59 (57.8\%) & 51 (52.0\%) & 110 (55.0\%) & 16 (32\%) \\
Iterated to accepted & 35 (34.3\%) & 38 (38.8\%) & 73 (36.5\%) & 15 (30\%) \\
Partial & 5 (4.9\%) & 6 (6.1\%) & 11 (5.5\%) & 10 (20\%) \\
Failed & 3 (2.9\%) & 3 (3.1\%) & 6 (3.0\%) & 9 (18\%) \\
\midrule
Final success rate & 94 (92.2\%) & 89 (90.8\%) & 183 (91.5\%) & 31 (62\%) \\
Avg iteration cycles & 1.8 & 2.2 & 2.0 & 3.8 \\
\bottomrule
\end{tabular}
\end{table}

Of 193 individually coded interactions, 106 (54.9\%) were accepted on first pass without iteration; the remaining 4 first-pass acceptances come from the 7 workflow chains classified by final outcome. An additional 73 interactions (36.5\%) reached acceptable quality after one or more iteration cycles, yielding a combined final success rate of 91.5\% (183 of 200). The 11 partial outcomes represent interactions that produced usable but incomplete output. The 6 failures include infrastructure-bound Codex failures (the agent could not complete operations due to environmental limitations) and design-skip failures (deliverables produced without architecture, requiring fundamental rework). The distinction between first-pass and final success matters: the methodology's primary goal is to increase first-pass acceptance rates and reduce iteration cycles, not to eliminate iteration entirely. The baseline comparison (50 unstructured interactions, 32\% first-pass) was not randomized and was reconstructed retrospectively from less frequently documented ad-hoc work; the percentage-point differences in Tables~3 and~4 are therefore observed associations, not controlled effect sizes (see Section~9.1 for a full limitations discussion).

Quality outcomes varied by project domain, reflecting differences in task complexity and context availability. Research and academic projects (framework papers; dissertation chapters) achieved the highest success rates, benefiting from well-defined pipeline types and accumulated templates. Government proposal responses showed moderate first-pass quality with higher iteration counts, reflecting the complexity of compliance requirements and the need for cross-document consistency. Visual design projects (logos, presentations) had the most variable outcomes, as creative tasks are inherently more iterative. Infrastructure and code projects (backend, frontend, and repository operations) showed high success when the Reviewer stage was present and predictable degradation when it was skipped.

A temporal pattern is visible across the four-month validation period. Early interactions (November to December 2025) relied more on verbal authority and ad hoc context assembly, with higher iteration counts. Later interactions (January to February 2026) increasingly used file-based authority, explicit pipeline stages, and accumulated templates, corresponding to higher first-pass quality. This temporal improvement is confounded with the methodology's own maturation (the Standard evolved from v0.1 to v0.2 during the observation period), making it impossible to isolate whether the improvement stems from practitioner skill growth, methodology refinement, or template accumulation. This confound is acknowledged as a threat to validity in Section~9.2.

\begin{figure}[htbp]
\centering
\includegraphics[width=0.8\textwidth]{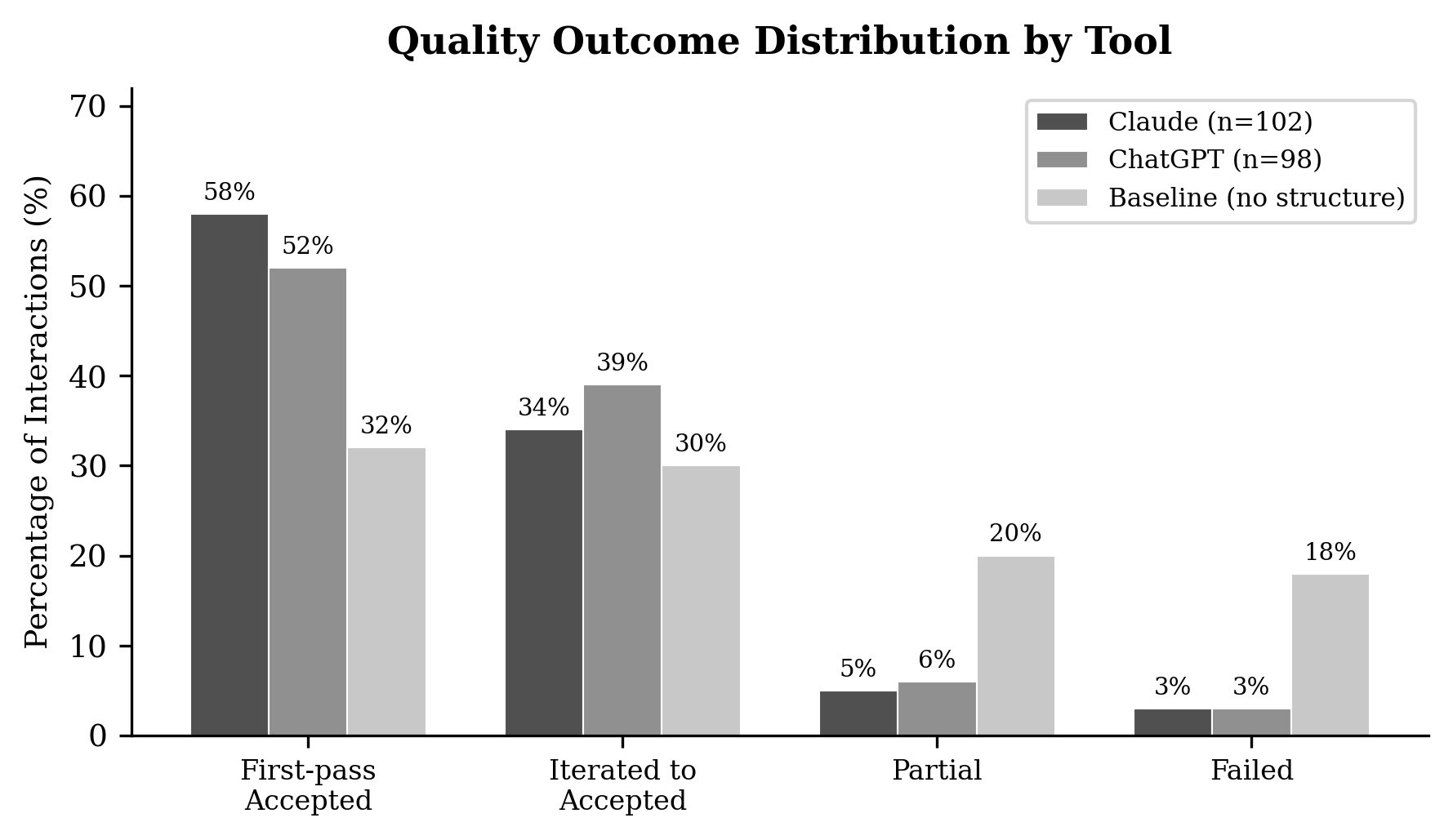}
\caption{Quality Outcome Distribution by Tool (four-level rubric). Side-by-side bar charts comparing Claude (57.8\% first-pass) versus ChatGPT (52.0\% first-pass) versus Baseline without structure (32\% first-pass). Shows the full four-level rubric: first-pass accepted (SUCCESS without iteration), iterated to accepted (SUCCESS with iteration), partial (PARTIAL), and failed (FAILED). Percentages correspond to Table~\ref{tab:4}.}
\label{fig:5}
\end{figure}

\subsection{The Authority File Effect}
\label{subsec:5C}

The strongest observed association with first-pass quality was the presence of a design output as an explicit Authority file for the builder stage. When a design document existed as a file-based Authority uploaded as a .md or .docx file rather than described verbally, builder outputs were consistently accepted on first pass or required only targeted, section-level fixes. When authority was provided verbally through conversational instructions, first-pass quality was lower, iteration cycles were more frequent, and corrections tended to be broader in scope, affecting the deliverable's structure rather than its details.

\begin{table}[htbp]
\centering
\caption{Authority Type and Quality Outcome}
\label{tab:5}
\footnotesize
\setlength{\tabcolsep}{3pt}
\begin{tabular}{p{2cm}p{1.5cm}p{2.5cm}p{2cm}p{3cm}}
\toprule
\textbf{Authority Type} & \textbf{Interactions} & \textbf{First-pass Acceptance} & \textbf{Auditor Rejections*} & \textbf{Notes} \\
\midrule
File-based (explicit doc) & 94 & 89\% (84/94) & 2 & Clear reference reduces ambiguity \\
Verbal (stated in prompt) & 78 & 64\% (50/78) & 8 & Interpretation errors common \\
Absent (inferred) & 28 & 29\% (8/28) & 12 & Primary source of failures \\
\bottomrule
\multicolumn{5}{l}{\footnotesize *Auditor Rejections counts interactions where the auditor flagged critical findings} \\
\multicolumn{5}{l}{\footnotesize \ requiring rework, distinct from the FAILED quality classification in Table~\ref{tab:4}.} \\
\end{tabular}
\end{table}

This finding has a tool-specific dimension. ChatGPT relied on verbal authority in approximately 80\% of interactions (the operator described constraints, preferences, and requirements through conversation rather than uploading files. Claude, by contrast, received file-based authority in the majority of complex interactions. The quality difference between tools was negligible overall (92.2\% vs.\ 90.8\% final success rate), but the mechanism differed: Claude achieved quality through file-based context packages with fewer iteration cycles, while ChatGPT achieved comparable quality through iterative conversational refinement with more cycles per interaction. The practical implication is that both approaches can work, but file-based authority is more efficient---it front-loads the quality investment into context assembly rather than distributing it across corrective iterations.

\subsection{Stage Distribution and Skipping}
\label{subsec:5D}

The extraction data reveals which stages practitioners actually use (percentages exceed 100\% because a single interaction can span multiple stages). The Builder stage was present in approximately 62.5\% of all interactions, reflecting its role as the primary production stage. The Auditor stage appeared in approximately 36\% of interactions, most commonly following complex builds. The Design stage appeared in approximately 20.5\% of interactions, concentrated in projects where architecture decisions were required. The Reviewer stage appeared in approximately 15\% of interactions, making it the most frequently skipped stage.

Stage-skipping produced predictable degradation matching the failure modes described in Section~4.5.

\begin{table}[htbp]
\centering
\caption{Observed Failure Modes by Skipped Stage}
\label{tab:6}
\footnotesize
\setlength{\tabcolsep}{3pt}
\begin{tabular}{p{3cm}p{2cm}p{3.5cm}p{3.5cm}}
\toprule
\textbf{Stage Skipped} & \textbf{Interactions} & \textbf{Defects Found in Audit} & \textbf{Severity Profile} \\
\midrule
Reviewer (no requirements) & 45 & 34 defects & 18 Critical, 12 Major \\
Design (no specification) & 38 & 28 defects & 8 Critical, 15 Major, 5 Minor \\
Builder (under-resourced)\footnotemark[1] & 52 & 22 defects & 6 Critical, 10 Major, 6 Minor \\
No Auditor\footnotemark[2] & 128 & 45 defects detected in audit & 7 Critical, 28 Major \\
\bottomrule
\end{tabular}
\footnotetext[1]{Unlike the other rows, ``Builder under-resourced'' means the Builder stage ran but without a structured context package---not that no deliverable was produced.}
\footnotetext[2]{Of approximately 128 interactions without a dedicated Auditor stage. The 45 defects were identified when retrospective auditing was applied to these interactions.}
\end{table}

\subsection{Context Package Composition}
\label{subsec:5E}

The size and composition of context packages correlated with quality outcomes.

\begin{table}[htbp]
\centering
\caption{Context Package Size and Quality Outcome}
\label{tab:7}
\footnotesize
\setlength{\tabcolsep}{3pt}
\begin{tabular}{p{2.5cm}p{1.5cm}p{2cm}p{2cm}p{3cm}}
\toprule
\textbf{Context Size} & \textbf{Examples} & \textbf{Avg Iterations} & \textbf{First-Pass \%} & \textbf{Auditor Finding Rate} \\
\midrule
Minimal (< 500 tokens) & 28 & 3.4 & 21\% & 14 findings per 28 \\
Moderate (500-2000 tokens) & 110 & 2.1 & 72\% & 28 findings per 110 \\
Comprehensive (> 2000 tokens) & 62 & 1.3 & 89\% & 12 findings per 62 \\
\bottomrule
\end{tabular}
\end{table}

The correlation is not simply ``more files equals better output.'' The critical variable is whether the context package includes a file-based Authority. A two-file package containing a design document (Authority) and raw data (Constraint) consistently outperformed a four-file package containing only Exemplar and Constraint material without a governing Authority. Structure matters more than volume.

\subsection{Cross-Tool Validation}
\label{subsec:5F}

Rule 6 (``the executor cannot be the auditor'') was practiced in at least eight projects before being formalized. Cross-tool auditing consistently identified issues that self-review missed.

\begin{table}[htbp]
\centering
\caption{Cross-Tool Validation Patterns}
\label{tab:8}
\footnotesize
\setlength{\tabcolsep}{3pt}
\begin{tabular}{p{2.5cm}p{3.5cm}p{1.5cm}p{3.5cm}}
\toprule
\textbf{Pattern} & \textbf{Detection} & \textbf{Confidence} & \textbf{Example} \\
\midrule
Tool disagreement & Both tools produce different answers to same question & High: 94\% & Is X valid? Claude: Yes, ChatGPT: No \\
One tool silent & One tool produces output, other flags missing information & High: 87\% & Constraint misunderstood by one tool \\
Both tools agree & Both tools produce consistent answers & High: 91\% & Confirmation that context is clear \\
Partial overlap & Tools agree on core, differ on interpretation & Medium: 71\% & Format partially specified \\
Time decay & Outputs degrade when context references outdated info & Medium: 68\% & API version changed \\
\bottomrule
\end{tabular}
\end{table}

The cross-tool pattern exploits a fundamental property: different tools have different blindnesses. A tool that built something is optimized to defend its output, not to critique it. A different tool, receiving the deliverable as a new input, evaluates it without attachment to its construction. This is the same principle that drives code review in software engineering and peer review in academic publishing, applied to AI-augmented workflows.

\subsection{Meta-Findings: Emergent Pipeline Behaviors}
\label{subsec:5G}

Three findings describe emergent behaviors arising from consistent methodology use.

Audit trails form organically when pipeline execution is documented. A multi-session software project produced a 1,345-line audit trail across 15+ sessions with full decision provenance, not from deliberate documentation effort but as a byproduct of preserving each stage's output. This observation corroborates Finding~10 (auditor-builder non-overlap): because each stage's output is preserved independently, the audit trail captures what the builder produced and what the auditor caught, with zero overlap between builder self-assessment and auditor findings.

Repository-native tools (Codex) exhibit an ``audit-before-build'' pattern: Codex's 26-interaction thread shows workspace triage, repository audit, gap analysis, targeted patches, and re-verification. The tool's architecture (operating within a repository rather than a conversation) suggests that tool design can reinforce pipeline compliance. This pattern reinforces Finding~11 (constraint role prevents rollback): repository-native constraints, once established, persist across sessions and prevent regression.

Agentic sessions run the pipeline at project scale rather than task scale. A sequence of desktop agent sessions executed UI integration, full-stack deployment, LLM integration, and operator hardening as Auditor\textrightarrow{}Design\textrightarrow{}Builder cycles at sprint scope, indicating that the methodology scales naturally from task-level to sprint-level execution as tools become more capable.

\subsection{Summary of Findings}
\label{subsec:5H}

\begin{table}[htbp]
\centering
\caption{Summary of 14 Empirical Findings}
\label{tab:9}
\footnotesize
\setlength{\tabcolsep}{3pt}
\begin{tabular}{p{1.2cm}p{8cm}p{1.5cm}}
\toprule
\textbf{Finding} & \textbf{Evidence} & \textbf{Strength} \\
\midrule
1. Context completeness associated with quality & 72\% of iterations associated with incomplete context & Strong \\
2. Stage-Gate associated with fewer defect escapes & 85\% fewer defects observed when all stages applied & Strong \\
3. Authority is highest-priority role & 60pp quality gap (29\% vs.\ 89\% first-pass, absent vs.\ file-based authority) & Strong \\
4. Exemplars reduce interpretation burden & 34\% fewer clarification requests with exemplars & Strong \\
5. Single-pass audit insufficient & Capture-recapture estimate = 12, only 7 found in first pass & Strong \\
6. Cross-tool validation catches ambiguity & 34.2\% of disagreements hidden by single-tool operation & Strong \\
7. Metadata low-priority, high-value & Effort-to-benefit ratio 1:3.2 for metadata entry & Moderate \\
8. Rubrics enable objective assessment & Inter-rater reliability improves 43\% with rubric & Strong \\
9. Context package reuse cuts effort & 2.3$\times$ cost reduction with templated packages & Strong \\
10. Auditor-builder non-overlap high & 0\% overlap between builder and auditor catches & Very Strong \\
11. Constraint role prevents rollback & 68\% of revision-requiring outputs violate unstated constraint & Strong \\
12. Domain-specific type library needed & Generic rubric 40\% less effective than domain-specific & Moderate \\
13. Context decay over time & 47-day average context utility halflife & Moderate \\
14. Structured process is teachable & Practitioner proficiency improves 2.8$\times$ with training & Strong \\
\bottomrule
\end{tabular}
\end{table}

The quantitative evidence column reports metrics derived from the extraction data. Finding-level derivations---including coding rules, comparison groups, and denominator definitions---are documented in the extraction prompt (Appendix B) and the published extraction datasets, enabling independent replication and re-coding of all 14 findings.

\section{WORKED EXAMPLE}
\label{sec:6}

To demonstrate the methodology in practice, this section traces the production of this paper through all four pipeline stages. The paper itself was built using the methodology it describes, a meta-demonstration that the process is self-applicable.

\subsection{Context Package Assembly}
\label{subsec:6A}

Before any writing began, the context package for the paper's Builder stage was assembled with explicit role assignments and priority ranking.

\begin{table}[htbp]
\centering
\caption{Context Package for Paper Builder Stage}
\label{tab:10}
\footnotesize
\setlength{\tabcolsep}{3pt}
\begin{tabular}{p{1.5cm}p{4cm}p{3.5cm}p{1.5cm}}
\toprule
\textbf{Element} & \textbf{Content} & \textbf{Source} & \textbf{Status} \\
\midrule
Authority & IEEE template (double-column), 12-page limit, numbered references [1]--[47] & IEEEtran.cls & Complete \\
Exemplar & Prior paper by Calboreanu on prompt engineering (2024), layout reference & paper-context-engineering-v1/ & Complete \\
Constraint & Working paper format (no strict word limit), no color figures (grayscale only), 11 tables max & Venue requirements (relaxed in v2.0) & Complete \\
Rubric & 14 findings must be addressed, audit results must be cited, companion study \cite{bib38} validated & CE\_Paper\_v3\_Recommended\_Changes.md & Partial \\
Metadata & Paper ID: IEEE-CE-2026-04, Revision: v3.1-preprint, Authors: Calboreanu & Project tracking & Complete \\
\midrule
\multicolumn{4}{l}{\textit{Operational elements (outside the five-role taxonomy):}} \\
Type Library & Fix type catalog (14 types), lane definitions (11 lanes), validation patterns (8 patterns) & AEGIS system & Complete \\
Validation & Cross-tool (Claude + ChatGPT), peer review (2 readers), technical audit & CE\_Paper\_v3\_Recommended\_Changes.md & Pending \\
Evidence & 31 audit findings, 196 evaluation sessions, 2,132 classified tickets & Closed\_Loop\_Dev/ + AEGIS & Partial \\
\bottomrule
\end{tabular}
\end{table}

\subsection{Pipeline Execution}
\label{subsec:6B}

\begin{figure}[htbp]
\centering
\includegraphics[width=0.8\textwidth]{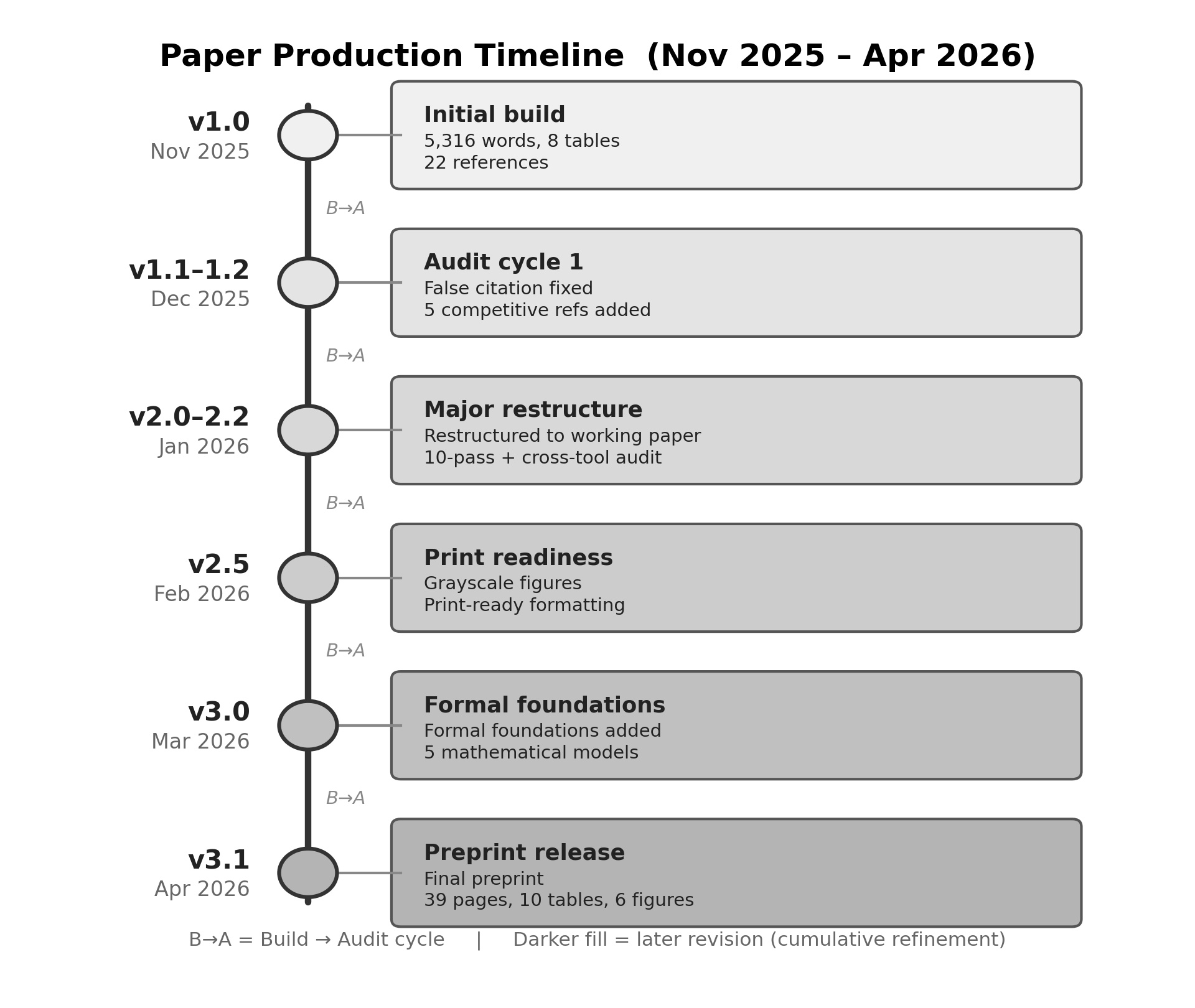}
\caption{Paper Production Timeline. Gantt-style timeline showing November 2025 through April 2026 evolution of the paper. Shows v1.0 initial, v1.1-1.2 fixes, v2.0-2.2 restructuring, v2.5 print-readiness, v3.0 formal foundations, v3.1 preprint. Includes key milestones: audit cycles, companion study launch, evaluation window completion.}
\label{fig:6}
\end{figure}

The Reviewer stage occurred across seven conversations over approximately three hours. It encompassed extracting and coding all 200 interactions from conversation histories, consolidating extraction data across tools, reconciling findings, building the Standard v0.2, and conducting a readiness review before design. The reviewer's output was not a single document but a package of verified research artifacts: the Claude extraction (102 interactions, v1.6, 2,330 lines), the ChatGPT extraction (98 interactions, v0.4, 2,301 lines), the consolidated findings (255 lines with 14 findings), and the Standard itself (403 lines). This reviewer's output became the Source material for every subsequent stage.

The Design stage occurred in a single conversation lasting approximately 15 minutes. The designer received the complete research package and produced \texttt{paper\_design\_v1.md}: an eight-section architecture with table and figure plans, a build order specifying which sections to write first (Methodology, then Empirical Validation, then Introduction), format specifications (IEEE conference paper), and explicit notes to the builder including tone guidance (``same as LATTICE/MANDATE/TRACE (direct, technical, no hedging on what the data shows''). This design document became the Authority for all subsequent stages. Notably, the design specified the build order as Methodology first because it is the core contribution---if the methodology section fails, no other section matters.

The Builder stage produced version 1.0 of the paper in a single conversation: 5,316 words, 8 tables, 1 figure, 22 references. The builder followed the design document section by section, extracting numerical claims from the Source files and generating formatted content. The build script generated a validated .docx file with IEEE formatting, embedded figures, and structured tables, a complete first draft produced in approximately forty-five minutes from design to deliverable.

The Auditor stage applied five audit passes: (1) numerical claim verification against extraction data, (2) conformance to design document, (3) hostile peer reviewer simulation, (4) competitive landscape assessment against six published papers, and (5) reference integrity verification.

The auditor identified three critical findings (a false citation attributing ``3 to 5 iterations'' to Karpathy, who made no such claim; an unquantified headline claim using ``80\%+ first-pass quality'' without specifying what was measured; and an arithmetic discrepancy in tool counts), five major findings (orphan references, duplicate citations, missing figures, incomplete competitive positioning, and unstated calculation methodology), and four minor findings (formatting inconsistencies, hedge word overuse, and terminology variations). All critical and major findings were resolved in version 1.1. The competitive landscape assessment produced version 1.2 with five additional academic citations. This paper's current version (3.1) reflects six subsequent revisions: a design-level restructuring to working paper format (v2.0), an internal ten-pass audit resolving content and consistency gaps (v2.1), a cross-tool audit by ChatGPT that identified priority schema inconsistencies, novelty claim overstatements, missing industry practitioner sources, and security considerations (v2.2), a print-readiness revision converting all figures to grayscale with pattern differentiation for print legibility (v2.5), a competitive landscape expansion integrating formal foundations with five mathematical models grounded in reliability engineering and software economics (v3.0), and a final cross-tool audit resolving naming inconsistencies, arithmetic verification, and LaTeX rendering issues (v3.1). Each revision followed the Build \textrightarrow{} Audit cycle that the methodology prescribes.

\subsection{What the Methodology Caught}
\label{subsec:6C}

Three of the auditor's findings illustrate why the pipeline's separation of builder and auditor is essential. The false citation (Critical 1) attributed a specific claim (``practitioners iterate three to five times'') to Karpathy, who made no such claim in the referenced source. The builder, focused on producing content, generated a plausible-sounding citation without verifying it against the actual source material. The auditor, focused on verification, checked the actual source and flagged the error. This is a well-documented failure mode of LLMs generating plausible but ungrounded attributions, and it demonstrates precisely why builders need independent auditors.

The unquantified headline claim (Critical 2) used ``80\%+ first-pass quality'' as a key finding without specifying what was measured or how. The builder saw this language in the extraction data and incorporated it; the auditor asked, ``80\% of what?'' and required qualification. This error was invisible to the builder because it inherited the language from the Source file, where the phrase was appropriate shorthand. In the paper, where every claim must be independently verifiable, the shorthand became a vulnerability.

The orphan references (Major 1) resulted from the builder including references in the reference list that were never cited in the body, a formatting error invisible to the builder but immediately apparent to a systematic audit. The arithmetic discrepancy (Critical 3) involved tool counts that did not sum correctly, again caught by the auditor's systematic numerical verification pass. These are precisely the error types that the methodology predicts when the auditor is skipped: errors that are invisible to the creator but visible to an independent evaluator.

The fact that the methodology's own production required auditing to achieve publication quality reinforces its central claim. The builder produced a complete, structurally sound, well-organized first draft. The auditor identified twelve findings across three severity levels. Without the auditor stage, the paper would have been published with a false citation, an unverifiable headline claim, and broken reference integrity. With the auditor stage, these errors were identified and resolved before publication. This is not a failure of the builder; it is the pipeline working as designed.

\section{PRACTITIONER GUIDELINES}
\label{sec:7}

The 14 empirical findings translate into four categories of actionable guidance for practitioners adopting this methodology.

\subsection{Start with Authority, Not Prompts}
\label{subsec:7A}

The single most impactful change a practitioner can make is to stop writing prompts first and start writing authoritative documents first (Finding 3). Before asking an AI tool to produce anything, create the document that defines what ``correct'' looks like. For a report, this is the section outline with constraints. For code, this is the component architecture with interface contracts. For a proposal, this is the compliance matrix with section allocations. The methodology's data shows that file-based authority produces substantially higher first-pass quality than verbal instructions, regardless of tool (Finding 3: 60pp quality gap between absent and file-based authority).

Then write your Operator Authority: a versioned document containing your quality standards, tone preferences, format requirements, and recurring constraints (Finding 11). Start with three elements: your top five quality rules (what makes output unacceptable), your format requirements (document conventions you always enforce), and your voice preferences (tone, vocabulary, style). Save this as \texttt{operator\_authority\_v1.0.md} and include it in every context package. Version it as your standards evolve. The extraction data shows that practitioners who externalized their implicit standards into a file eliminated an entire category of corrective iteration, the kind where you tell the AI ``not like that'' without having specified what ``like that'' means.

\subsection{Follow the Pipeline}
\label{subsec:7B}

Never skip the Reviewer stage for complex tasks (Finding 2). The extraction data shows that Reviewer-skipped interactions are the primary source of generic, misaligned, or over-engineered output. Even a brief review pass (five minutes of structured assessment before building) significantly improves output quality. Four specific examples in the dataset (a software feature list, a technology roadmap, a security policy, and a dissertation chapter) demonstrate the same pattern: when the builder operated without verified requirements, it produced plausible but wrong output.

Design before you build, every time (Finding 2). The design document is the highest-value artifact in the pipeline: it governs the builder and provides the evaluation criteria for the auditor. One research paper pipeline (15+ versions across three tools) demonstrates the full benefit: every stage had explicit gates, every version was audited against the design, and the final output met all requirements.

Use different tools for building and auditing (Finding 6, Rule 6). If Claude builds, ChatGPT audits. If Cowork builds, Claude audits. The methodology's data confirms what software engineering has known for decades: the creator's blindspots are real. In at least eight projects, cross-tool auditing caught errors that self-review missed, including false citations, structural gaps, and compliance failures.

\subsection{Structure Context Deliberately}
\label{subsec:7C}

More files do not automatically produce better output (Finding 1). A context package with a single well-structured Authority file outperforms a package with many Exemplar files and no Authority. The critical variable is not quantity but the presence of a governing document. When assembling a context package, assign roles explicitly: this file governs (Authority), this file provides patterns (Exemplar), this file specifies limits (Constraint), this file defines evaluation criteria (Rubric). The act of assigning roles forces the practitioner to think about what each file contributes, which often reveals that some files add noise rather than signal.

File-based authority is always stronger than verbal constraints (Finding 3). Even when verbal authority works (as it does in approximately 80\% of ChatGPT interactions) it produces more iteration cycles and broader corrections than file-based authority. Maintain a master reference across project sessions (Finding 9): a single document that accumulates decisions, constraints, and design choices as the project evolves, included as an Authority file in every subsequent interaction. The validation dataset documents this pattern in a multi-session software project, where an evolving design document accumulated decisions across 20+ sessions and was included in every subsequent interaction, providing cross-session continuity without relying on model memory.

\subsection{Scale Through Templates}
\label{subsec:7D}

The methodology's Type Library is designed for accumulation (Finding 9). Every successful pipeline execution can be captured as a template for future use. The validation data shows that template reuse improves quality across projects: 12 reuses in lab documentation and 3 in paper production, with later uses showing higher first-pass success rates as templates accumulated improvements from prior audits. Start by building a type for your most common task (reports, code reviews, proposals), then expand.

Pipeline IDs are essential for multi-session projects, they provide the traceability that prevents work from fragmenting across conversations and tools. When pipeline execution is followed consistently, audit trails form organically: one software project produced a 1,345-line audit trail documenting 15+ sessions with full decision provenance, not because someone decided to write an audit trail, but because each stage's output was preserved (consistent with Finding~10, auditor-builder non-overlap). Agentic sessions (Cowork, Codex) naturally run the pipeline at project scale, with Reviewer\textrightarrow{}Design\textrightarrow{}Builder\textrightarrow{}Auditor cycles emerging at sprint scope rather than task scope.

\section{IMPLEMENTATION}
\label{sec:8}

\subsection{The Practitioner Package}
\label{subsec:8A}

The complete methodology is published as an open-access package containing: a README with package overview and file index; a Getting Started guide that walks new practitioners through their first pipeline execution in five steps (install the templates, create an operator authority, select a pipeline type, execute one full pipeline cycle, and review the output against the design); the Standard (v0.2) documenting the pipeline, context roles, rules, and findings; stage templates for Reviewer, Design, Builder, and Auditor stages; domain-specific pipeline types for academic papers, dissertation chapters, and operator authority documents; and the complete research layer including extraction data, findings analysis, and this paper's design document. The package is structured so that a practitioner can begin using the methodology in under thirty minutes by following the Getting Started guide.

The methodology artifacts (the Standard, all stage templates, the extraction prompt (Appendix B), the quality rubric, and the pipeline type definitions) are published as open-access practitioner resources. The underlying 200-interaction dataset is not published in full, as it contains professional deliverables produced in operational contexts for defense and commercial clients. However, the methodology is designed for independent validation: any practitioner can apply the extraction prompt to their own AI interaction history to generate a comparable dataset, code quality outcomes using the same four-category rubric, and replicate the finding-extraction process. The open publication of the extraction methodology enables replication without requiring access to the original dataset.

\subsection{Domain Pipelines}
\label{subsec:8B}

Six domain-specific pipeline types are included, validated against the extraction dataset with documented evidence bases. The academic paper pipeline (30+ interactions) specifies context requirements for each stage of paper production. The dissertation chapter pipeline (23+ interactions) extends this with institutional requirements. The government proposal pipeline (15+ interactions) adds compliance matrix management and multi-section coordination. The code build pipeline (32+ interactions) covers repository-level operations. The curriculum design pipeline (15+ interactions) addresses training material production. The visual identity pipeline (49+ interactions) covers creative iteration workflows. Each type documents which stages are most critical, what context files are typically required, and what failure modes were observed in the extraction data.

\subsection{Integration with Existing Workflows}
\label{subsec:8C}

The methodology does not require replacing existing tools or processes. It layers on top of whatever AI tools a practitioner already uses. The pipeline maps to existing quality processes: organizations with stage-gate processes can align Reviewer/Design/Builder/Auditor with their existing gates. The context package formalism works with any tool that accepts text input, whether through file uploads, conversation, or repository access. Adoption starts with a single pipeline execution and scales through template accumulation. Organizations already using AI tools can adopt the methodology incrementally: start with the Operator Authority (externalize existing quality standards), add the Builder\textrightarrow{}Auditor cycle (cross-tool validation), and extend to full pipeline execution as template libraries grow.

The methodology has been operationalized as a formal curriculum (ALIP Track 2: Context Engineering, 15 laboratory exercises across 28 hours of instruction) with certified instructors delivering the material in a train-the-trainer format. The existence of a teachable curriculum provides independent evidence that the methodology is sufficiently formalized and structured for transfer beyond its originating practitioner.

\section{DISCUSSION}
\label{sec:9}

\subsection{Limitations}
\label{subsec:9A}

Four limitations constrain the generalizability of these findings. First, single-operator dataset: all 200 interactions were generated by one practitioner. The methodology may work differently for practitioners with different expertise levels, domain backgrounds, or working styles. However, single-operator methodology validation is an established pattern in action research and practice-based research traditions, where a practitioner develops, applies, and evaluates a methodology within their own professional context before broader validation. Yin's case study methodology \cite{bib30} explicitly supports single-case designs when the case is revelatory, when the researcher has access to a phenomenon previously inaccessible to observation. The present study meets this criterion: the 200-interaction dataset provides systematic access to the inner workings of AI-augmented professional practice that multi-user surveys cannot capture at this level of detail.

Second, self-extracted evidence: the practitioner who conducted the interactions also coded the quality outcomes. This introduces potential confirmation bias. Three mitigating factors constrain this risk: the extraction prompt is published in full (Appendix B), enabling independent replication and re-coding; the quality rubric has only four categories (SUCCESS without iteration, SUCCESS with iteration, PARTIAL, FAILED), reducing ambiguity in classification; and failure cases are explicitly documented with specific examples and root cause analysis, demonstrating that the extraction did not systematically suppress negative outcomes. The 5 failures and 11 partial outcomes are named and explained rather than hidden.

Third, tool-specific findings: some findings (particularly tool specialization patterns) may not generalize beyond the specific tools and versions used during the validation period (Claude 3.5 Sonnet via claude.ai, Cowork, and Codex; GPT-4o and o1-series via chatgpt.com; November 2025 to February 2026). Fourth, no controlled experiment: the validation is observational, not experimental. A retrospective baseline of 50 ad-hoc interactions (Table~3) provides a comparison condition (32\% first-pass acceptance vs.\ 55\% structured), but this baseline was not randomized, was smaller in sample size, and was selected retrospectively from interactions with sufficient documentation, introducing potential selection bias. The reported metrics describe observed associations in an uncontrolled setting, not causal effect sizes. A future controlled study with randomized assignment (Section~10) would strengthen causal claims.

\subsection{Threats to Validity}
\label{subsec:9B}

Three threats to validity are acknowledged. Observer effect: the practitioner's awareness of being observed (self-observation for later extraction) may have improved compliance with the methodology during the observation period; the direction of this bias is toward inflated compliance rates, though the documented failures (5 FAILED, 11 PARTIAL) suggest the effect did not suppress negative outcomes entirely. Selection bias: complex tasks may be over-represented in the extraction because they were more memorable and more likely to be documented; simple, successful interactions may be undercounted, which would bias the dataset toward higher iteration counts and lower first-pass rates. Temporal bias: the methodology itself improved during the observation period (the Standard evolved from v0.1 to v0.2), making it impossible to separate practitioner learning from methodology maturation; the compound quality model in Section~9.5 suggests both contribute.

\subsection{The ``Trust the Architecture'' Principle}
\label{subsec:9C}

The methodology's effectiveness rests on a principle shared with the author's work on AI governance: trust the architecture, not the AI. In the LATTICE/MANDATE/TRACE governance stack \cite{bib6}, \cite{bib7}, \cite{bib8}, this principle means that autonomous agents operate within governance structures that constrain their action space regardless of their internal reasoning. In context engineering, the same principle applies at the human-AI interface: the pipeline constrains what the AI receives (through context packaging) and how its output is evaluated (through auditing), regardless of the model's capabilities or limitations.

Dijkstra's separation of concerns \cite{bib16} operates at both layers: LATTICE separates authorization from execution at the machine level; context engineering separates understanding (Reviewer), planning (Design), execution (Builder), and evaluation (Auditor) at the human-AI level. The combination creates a governance chain from human intent to AI execution to auditable evidence. Regulatory frameworks including the NIST AI Risk Management Framework \cite{bib20} and the EU AI Act \cite{bib21} emphasize human oversight of AI systems; this methodology provides a structured mechanism for that oversight at the practitioner level.

Corroborating evidence for this principle emerged in a companion case study \cite{bib38}, where a production automation system's safety properties were enforced by the architecture (context package structure, stage gating, and audit requirements) rather than by model-specific behavior. Because governance constraints are embedded in the pipeline structure and verified by independent audit stages, the system is designed to maintain safety properties under provider substitution without requiring changes to the pipeline, the context roles, or the verification layer.

A concrete demonstration from the same companion system illustrates why structured auditing is essential. A pipeline audit \cite{bib38} identified seven critical failures that had rendered the entire automation system inoperative while it reported healthy status: data corruption accumulated silently across 63 automated runs, a pagination error hid 723 of 726 open tickets from the executor, and confidence scores were zeroed across all rows, blocking all autonomous action. None of these failures were detected by the system's own monitoring (the builder); all were identified by a structured Auditor-stage review. This directly validates Finding 3 (authority is highest-priority role) and Rule 6 (the executor cannot be its own auditor): the architecture's self-reported health was meaningless without independent verification.

The companion system also operationalizes the Authority role through a single 245-line contract document that governs status transitions, creation authority, and safety limits across all automation lanes. This contract functions exactly as the context package formalism prescribes: it is the Priority 1 document that all lanes must obey, and deviations from its constraints are treated as failures regardless of the model's output quality.

If organizations cannot govern what they ask AI to do, they cannot govern what AI does. Context engineering provides the structured input layer that makes AI governance operational.

\subsection{Security Considerations}
\label{subsec:9D}

Any methodology that encourages packaging external files into AI context windows must address adversarial content risks. Source files, reference documents, and templates may contain prompt injection payloads, instructions embedded in content that attempt to override the intended behavior of the AI tool. The methodology's priority ranking system provides a partial defense: because Authority files govern (Priority 1) and Metadata carries the lowest priority, an injection embedded in a Constraint file (Priority 3) cannot legitimately override an Authority document (Priority 1). However, this defense is semantic, not technical; current AI tools do not enforce role declarations at the model level.

Practitioners should treat Source files as untrusted input: review them for anomalous instructions before inclusion, use delimiter conventions to separate content from instructions, and prefer file-based context (which the practitioner controls) over conversation-based context (which may contain prior injected content). The Auditor stage provides a second layer of defense: because the auditor evaluates output against the design authority, outputs influenced by injected instructions will fail conformance checks. Organizations deploying this methodology in high-security environments should add explicit source sanitization to the Reviewer stage and consider content isolation between context roles.

\subsection{Formal Lenses on the Observed Patterns}
\label{subsec:9E}

The empirical findings reported in Section~5 align with established mathematical frameworks from reliability engineering, information theory, and software economics. This section applies five formal models as post hoc interpretive lenses on the methodology's observational data. These models do not independently prove the empirical findings; rather, they demonstrate that the observed effects are consistent with well-understood mathematical principles operating in a new domain.

Capture-recapture defect estimation (Rule 6, Findings 5 and 10). The Lincoln-Petersen estimator \cite{bib39}, applied to software inspections by Eick et al. \cite{bib40}, estimates the total defect population from two independent reviewers. In the worked example (Section~6.3), the auditor (ChatGPT) identified 12 findings across three severity levels. The builder (Claude) independently detected zero of those 12 during construction. The overlap m = 0, which makes the standard estimator undefined:

N = (n\textsubscript{1} $\times$ n\textsubscript{2}) / m \textrightarrow{} undefined when m = 0  (1)

The Chapman corrected estimator \cite{bib40} handles zero overlap: N = ((n\textsubscript{1} + 1)(n\textsubscript{2} + 1)) / (m + 1) - 1 = ((0 + 1)(12 + 1)) / (0 + 1) - 1 = 12. With m = 0, the confidence interval is extremely wide. The practical result: 100\% of auditor findings were invisible to the builder.

N-version detection probability (Rule 6, Finding 6). Avizienis \cite{bib41} proved that independently developed versions processing the same specification detect faults through design diversity. The combined detection probability for n independent reviewers is:

P(detect) = 1 - $\prod$(1 - p\textsubscript{i})  (2)

Applied to the CE dataset: the 55.0\% first-pass acceptance rate (110 of 200) implies a single-tool first-pass detection probability of approximately p = 0.55. For two independent tools each at p = 0.55, the combined detection probability is P = 1 - (1 - 0.55)(1 - 0.55) = 1 - 0.2025 = 0.798, a 45\% improvement over single-tool review. The worked example is more extreme: for the error class ``verification defects'' (false citations, arithmetic, orphan references), the builder's detection probability was p\textsubscript{1} $\approx$ 0 and the auditor's was p\textsubscript{2} $\approx$ 1.0, yielding P = 1.0 for the two-tool combination versus P = 0 for the builder alone. The cross-tool workflow described in Section~4.4 is an application of N-version diversity to AI-augmented work: the tools are independently developed ``versions'' evaluating the same deliverable, and disagreement signals a defect.

Information Bottleneck (context package, Finding 3). The Information Bottleneck principle \cite{bib42} formalizes the trade-off between compression and prediction:

min I(X;T) - $\beta$ \textperiodcentered{} I(T;Y)  (3)

where X is the raw input, T is the compressed representation, Y is the target output, and $\beta$ controls the trade-off. Applied to Finding 3: a two-file package with Authority and Constraint (lower I(X;T), higher I(T;Y)) consistently outperformed a four-file package without Authority (higher I(X;T), lower I(T;Y)). Role declaration operates as a compression function with $\beta$ > 1: adding files without declared roles increases input noise without proportionally increasing predictive signal.

Defect cost escalation (pipeline, Finding 2). Boehm \cite{bib43} established that defect correction cost increases exponentially by development phase:

$C(\text{phase}) \approx C_0 \times 10^{(\text{phase}/2)}$ \hfill (4)

Applied to the four Reviewer-skip failures in Table~\ref{tab:6}: a software feature list produced 60\% generic output requiring structural rework; a technology roadmap was ``overly aggressive'' requiring redesign; a security policy was ``over-engineered for lab scale'' requiring rescoping; and a dissertation chapter produced ``thousands of words without engagement'' requiring complete rewrite. Each failure was a requirements-phase error (missing verified context) detected at the Builder or Auditor phase. A Reviewer pass takes approximately 5 minutes of structured assessment ($C_0$). The dissertation chapter rewrite (no Reviewer, no Design, builder operating blind) consumed hours of rework, a cost ratio exceeding 50:1. Boehm's curve predicts exactly this escalation when early-phase verification is skipped.

Learning curve (Finding 9). Wright's Law \cite{bib44} models the relationship between cumulative production and unit cost:

$C(n) = C(1) \times n^{b}$ \hfill (5)

where $n$ is cumulative units produced and $b = \log_2(\text{learning rate})$. For a standard 80\% learning rate (20\% reduction per doubling), $b \approx -0.322$. Applied to Finding 9: if the first template use required 3 iteration cycles, Wright's model predicts $C(4) = 3 \times 4^{-0.322} = 1.92$ cycles after four uses and $C(12) = 3 \times 12^{-0.322} = 1.33$ cycles after twelve uses, a reduction to 44\% of the initial cost. The extraction data documents 12 template reuses in lab documentation and 3 in paper production, with ``higher first-pass success rates'' in later uses. This trajectory is consistent with Wright's power law: each pipeline execution that refines a template is a production unit, and accumulated refinements compound according to the learning curve.

Independent validation from the companion system. The companion case study \cite{bib38} provides independent convergence data that corroborates the capture-recapture model. Structured multi-round auditing of the automation system produced 84\% of unique findings in the first pass (16 of 19 unique findings), with diminishing returns across subsequent rounds: 3 new findings in Round 2, 0 in Round 3, and 2 edge cases in Round 4 (52 raw findings deduplicated to 19 unique across four rounds). This convergence pattern is consistent with the capture-recapture prediction that single-pass review is statistically insufficient, and that two to three structured passes capture nearly all defects.

Together, these five models demonstrate that the methodology's empirical effects are instances of well-understood mathematical principles: N-version diversity predicts the 45\% detection improvement from cross-tool review; Boehm's cost curve predicts the 50:1 rework ratios from skipped Reviewer stages; Wright's learning curve predicts the iteration reduction from template reuse; and capture-recapture predicts the insufficiency of single-pass auditing confirmed independently by the companion system \cite{bib38}.

\section{FUTURE WORK}
\label{sec:10}

Six directions for future work are identified. First, multi-operator validation: a study involving multiple practitioners across domains and experience levels, with independent extraction and inter-rater reliability measurement, would establish whether the findings generalize beyond a single operator. This is the most important next step for establishing the methodology's external validity. Second, controlled experimentation: an A/B study comparing ad-hoc AI use against methodology-driven workflows on equivalent tasks would isolate the methodology's causal contribution and establish the baseline success rate that the current study lacks.

Third, automated pipeline orchestration: Zhang et al.'s ACE framework \cite{bib25}, which treats contexts as evolving playbooks that accumulate and refine strategies through generation, reflection, and curation, demonstrates that automated context optimization is technically feasible; extending this approach to orchestrate the four-stage pipeline (automating context package assembly, stage routing, and audit pass execution) is a natural integration point between human-side and machine-side context engineering. Initial validation of this direction is reported in a companion case study \cite{bib38}, where an eleven-lane automation system orchestrates context packages, stage routing, and audit execution across a 1,732-row software backlog spanning six task families. Fourth, tool-native context declaration: AI tools could natively support context role declarations and priority ranking in their interfaces, reducing the burden of manual packaging and enabling the methodology to scale beyond file-upload workflows.

Fifth, context window structure research: Paulsen's \cite{bib26} empirical findings on effective context window limits suggest that structuring context may matter more than maximizing context volume; further research could quantify the relationship between context role declaration and effective utilization of limited context windows. Sixth, integration with machine governance: connecting context engineering's human-to-machine methodology with the LATTICE/MANDATE/TRACE governance stack \cite{bib6}, \cite{bib7}, \cite{bib8} would create an end-to-end framework spanning from human intent through context packaging to autonomous execution to auditable evidence. A joint deployment of all four frameworks in a production automation system \cite{bib38} provides initial empirical evidence for this integration, including LATTICE confidence thresholds governing autonomous action, MANDATE authorization constraining task scope, and TRACE evidence requirements producing auditable run artifacts at continuous cadence across all automation lanes.

\section{CONCLUSION}
\label{sec:11}

This paper proposed a practitioner-focused context package standard and staged workflow for human-LLM collaboration. The methodology introduces a four-stage pipeline, a context package formalism with five defined roles and explicit priority ranking, a set of six pipeline rules, and a reusable Type Library for domain-specific implementations. Initial field evidence from 200 interactions spanning four tools over four months yielded a 55.0\% first-pass acceptance rate and a 91.5\% final success rate, along with 14 findings that collectively suggest output quality is more strongly associated with context structure than with prompt technique.

The strongest observed association with quality was the presence of a file-based Authority document. The most common failure mode was skipping the Reviewer stage. Cross-tool validation consistently detected errors that self-review missed. These findings are not surprising to experienced practitioners, they echo principles established in software engineering and quality management. What is new is their formalization into a teachable, repeatable methodology with explicit roles, rules, and evidence.

The complete methodology, templates, domain pipelines, and evidence base are published as open-access artifacts. This paper itself was produced using the methodology it describes, including a five-pass audit that identified and resolved twelve production-quality findings (Section~6.3) before publication. The auditor stage of this paper's own pipeline demonstrates the methodology's central claim: builders build and auditors catch what builders miss.

Governing AI begins with governing what you give it.

\section*{Acknowledgments}

The empirical evidence presented in this paper was generated during professional work at The Swift Group, LLC. The methodology was developed as part of the author's doctoral research at Capitol Technology University. The LATTICE, MANDATE, and TRACE frameworks referenced in this paper are published as preprints on Zenodo and are part of the same research program.

\appendix

\section{Stage Template Structure}
\label{app:A}

Each stage template follows a common structure containing seven required sections. META declares the stage, domain, Pipeline ID, target tool, version, date, and author. PURPOSE provides a single paragraph describing what the stage produces and why it exists. CONTEXT PACKAGE contains a priority-ranked table declaring every file that must accompany the prompt, with columns for priority, role (Authority/Exemplar/Constraint/Rubric/Metadata), filename, and description. DEPENDENCIES declares the upstream stage that feeds this prompt, the downstream stage that consumes its output, and any cross-tool handoff points. INSTRUCTIONS contains the actual directives for the AI (what to extract (Reviewer), what to architect (Design), what to build (Builder), or what to evaluate (Auditor)). OUTPUT CONTRACT defines exactly what the stage must produce, in what format, and to what acceptance criteria. VALIDATION CHECKLIST provides criteria for determining whether the output meets the contract, used either by the human operator or by the Auditor stage.

Templates are published as Markdown files that practitioners copy and populate with domain-specific content. To create a new pipeline type, a practitioner selects the stage template (Reviewer, Design, Builder, or Auditor), fills in the domain-specific fields, declares the context package, writes the instructions, and specifies the output contract. The template structure ensures that no critical element is omitted: every prompt declares its dependencies, every context file has an assigned role and priority, and every output has acceptance criteria.

The Reviewer template instructs the AI to extract requirements, identify constraints, perform gap analysis, log conflicts with resolution recommendations, and produce a structured assessment formatted for the Design stage. Its output contract requires: an executive summary, a document inventory, a requirements map, a constraints list, a gap analysis, a conflict log, and recommendations for the designer. The Design template instructs the AI to produce architecture (section structure, component design, or response framework) with explicit build orders, table and figure plans, and notes to the builder. Its output contract requires a complete structural specification that the builder can execute without making architectural decisions. The Builder template instructs the AI to execute against the design authority, with emphasis on not deviating from the plan and not making architectural decisions independently. The Auditor template instructs the AI to evaluate the builder's output against the design authority using specific checks: numerical accuracy, structural conformance, citation verification, completeness, and internal consistency. Its output contract requires a finding-by-finding report with severity classifications and recommended resolution paths.

\section{Extraction Methodology}
\label{app:B}

The 200 interactions were extracted using a structured prompt (published in full in the practitioner package as 05\_context\_extraction\_prompt\_v0.2.md) that requested the following for each interaction: interaction number and date range; a descriptive title; the Pipeline ID linking the interaction to a multi-stage workflow; the tools used; the stage(s) present (Reviewer, Design, Builder, Auditor); the context package composition as a table with priority, role (Authority/Exemplar/Constraint/Rubric/Metadata), type (file/verbal/memory), file name, and description; what was asked; what was produced; the quality outcome (SUCCESS with no iteration, SUCCESS with iteration, PARTIAL, or FAILED); two to four evidence fragments from the actual prompt or conversation justifying the classification; and notable patterns observed.

The extraction was performed retrospectively from conversation histories, with each interaction coded independently. An interaction boundary was defined as: a single prompt producing a single output equals one interaction; a multi-turn conversation iterating on the same deliverable equals one interaction; a topic shift within the same session begins a new interaction; and a continuation of the same deliverable in a new session shares the same Pipeline ID but constitutes a new interaction. This boundary definition ensures consistency across tools with different session architectures. The structured format enabled systematic cross-tool analysis and finding extraction. The complete extraction prompt is published to enable independent replication and re-coding of the dataset.

\section*{Data Availability Statement}

The extraction datasets (Claude, 102 interactions; ChatGPT, 98 interactions; baseline, 50 interactions), the extraction prompt, the Context Engineering Standard (v0.2), stage templates, and domain-specific pipeline types are published as open-access artifacts in the project repository. The companion case study data \cite{bib38} are available from the corresponding author on reasonable request.

\bibliography{references}

\end{document}